\documentclass{article}

\usepackage{microtype}
\usepackage{graphicx}
\usepackage{booktabs} 

\usepackage{url}            
\usepackage{amsfonts}       
\usepackage{nicefrac}       
\usepackage[labelfont=bf]{caption}
\usepackage{subcaption}
\usepackage{amsmath, amsthm}

\usepackage{wrapfig}
\usepackage{xcolor}
\usepackage{xspace}
\usepackage{enumitem}

\usepackage[firstpage]{draftwatermark}
\SetWatermarkText{
 \hspace*{3.5in}
 \raisebox{10.0in}{
  \includegraphics[height=0.9in]{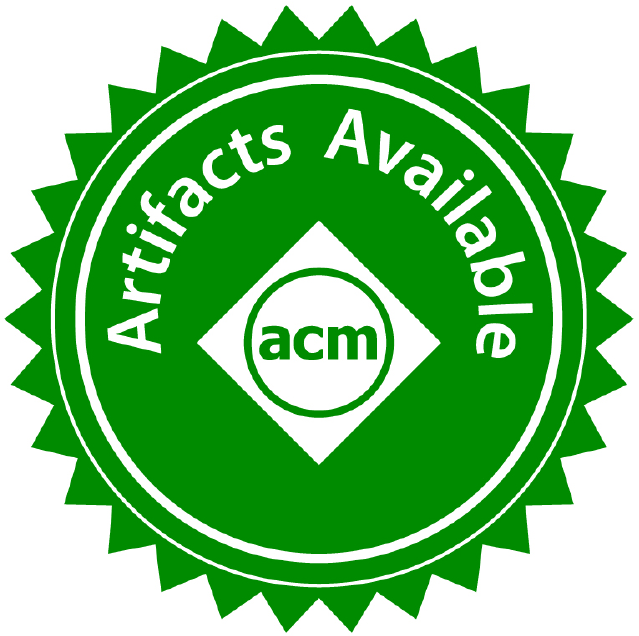}
  \includegraphics[height=0.9in]{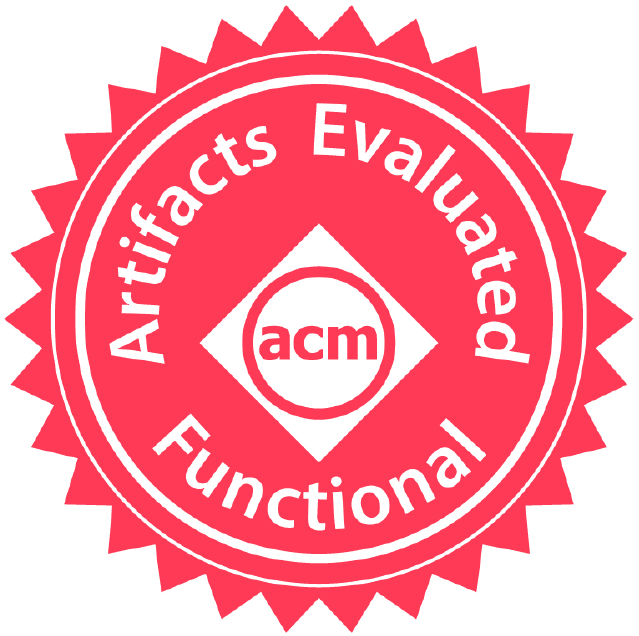}
 }
}
\SetWatermarkAngle{0}

\theoremstyle{plain}

\newcommand{\eps}{\varepsilon}

\usepackage{hyperref}

\usepackage[accepted]{mlsys2023}

\mlsystitlerunning{On Noisy Evaluation in Federated Hyperparameter Tuning}

\begin{document}

\twocolumn[
\mlsystitle{On Noisy Evaluation in Federated Hyperparameter Tuning}




\begin{mlsysauthorlist}
\mlsysauthor{Kevin Kuo}{csd}
\mlsysauthor{Pratiksha Thaker}{mld}
\mlsysauthor{Mikhail Khodak}{csd}
\mlsysauthor{John Nguyen}{meta}
\mlsysauthor{Daniel Jiang}{meta}
\mlsysauthor{Ameet Talwalkar}{mld}
\mlsysauthor{Virginia Smith}{mld}
\end{mlsysauthorlist}

\mlsysaffiliation{csd}{Computer Science Department, Carnegie Mellon University}
\mlsysaffiliation{mld}{Machine Learning Department, Carnegie Mellon University}
\mlsysaffiliation{meta}{Meta AI}

\mlsyscorrespondingauthor{Kevin Kuo}{kkuo2@andrew.cmu.edu}

\mlsyskeywords{Machine Learning, MLSys}

\vskip 0.3in

\begin{abstract}
\vspace{.05in}
Hyperparameter tuning is critical to the success of federated learning applications. Unfortunately, appropriately selecting hyperparameters is challenging in federated networks, as issues of scale, privacy, and heterogeneity introduce noise in the tuning process and make it difficult to faithfully evaluate the performance of various hyperparameters. In this work we perform the first systematic study on the effect of noisy evaluation in federated hyperparameter tuning. We first identify and rigorously explore key sources of noise, including client subsampling, data and systems heterogeneity, and data privacy. Surprisingly, our results indicate that even small amounts of noise can have a significant impact on  tuning methods---reducing the performance of state-of-the-art approaches  to that of naive baselines. To address noisy evaluation in such scenarios, we propose a simple and effective approach that leverages public proxy data to boost evaluation signal. Our work establishes general challenges, baselines, and best practices for future work in federated hyperparameter tuning.
\end{abstract}
]



\printAffiliationsAndNotice{}  

\section{Introduction}
\vspace{-.05in}
Hyperparameter tuning---the process of selecting the hyperparameters of a learning algorithm---is crucial for achieving high-performing models in machine learning. Hyperparameter (HP) tuning is essential for cross-device federated learning (FL) applications, which consider training machine learning models over large heterogeneous networks of devices such as mobile phones or wearables~\citep{mcmahan_communication-efficient_2017}. Although FL methods often rely on additional hyperparameters~\citep{li_federated_2020-1,reddi_adaptive_2020,charles_large-cohort_2021}, the budget for tuning such parameters may be particularly small due to computational and privacy-related constraints. Developing methods for federated HP tuning has thus been identified as a critical area of research~\citep{kairouz_advances_2021,khodak_federated_2021}.


Unfortunately, federated networks introduce the additional challenge of \textit{noisy evaluation}, which can prevent HP tuning methods from properly evaluating HP performance. A clear source of evaluation noise arises from client subsampling. As data is distributed across potentially millions of intermittently available clients, HP tuning algorithms must rely on signals from only a small subset of a much larger validation population~\citep{bonawitz_towards_2019}. 

\begin{figure}[!h]
    \centering
    \includegraphics[width=7cm]{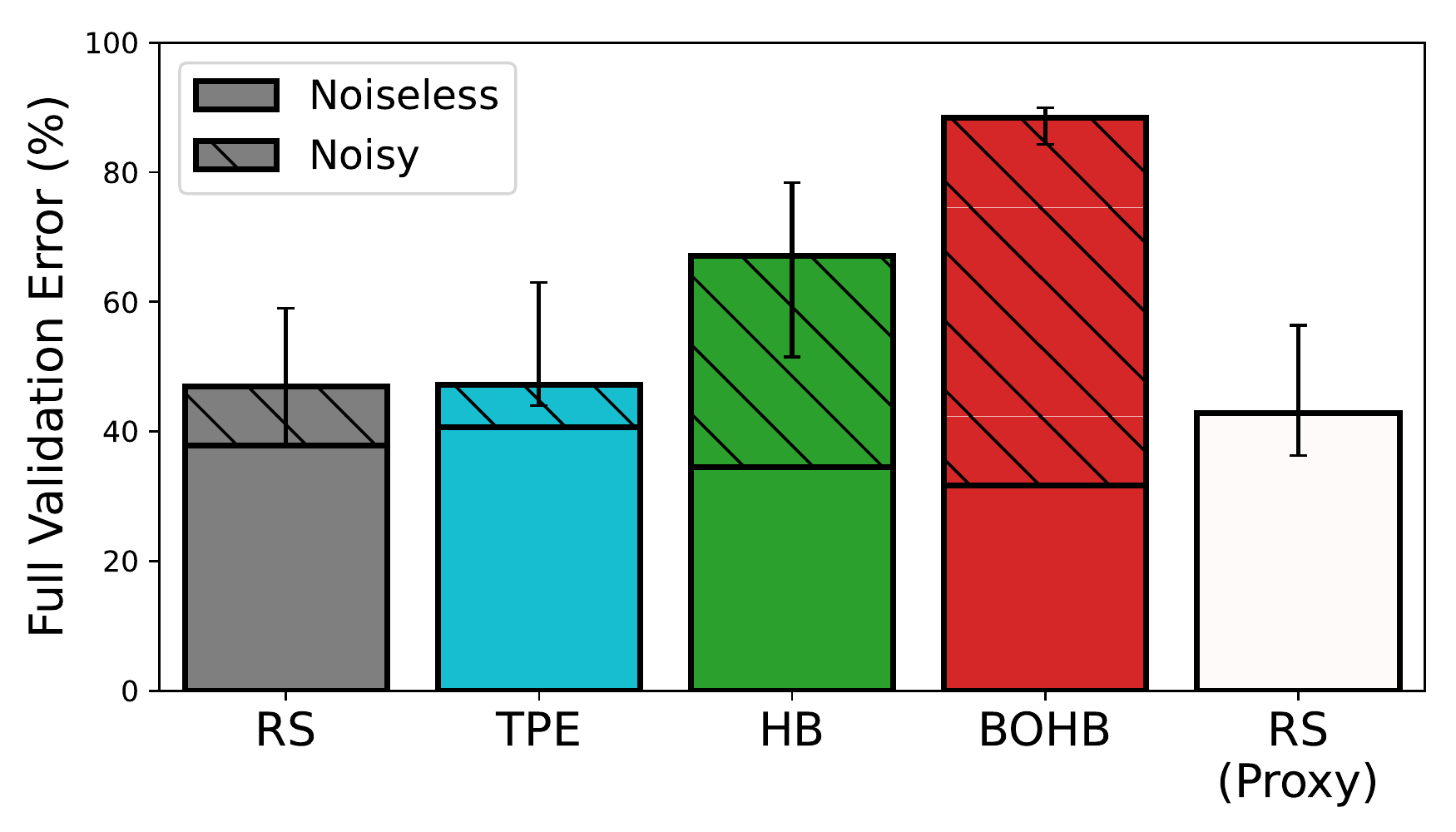}
    \vspace{-.1in}
    \caption{Error (lower is better) of a CIFAR10 model found by various HP tuning methods. With FL noise, more complex methods underperform relative to simple random search (RS). RS on proxy data is a strong baseline which is unaffected by noisy validation data. Bars only show performance at 1/3rd of the tuning budget to highlight faster convergence of HB and BOHB in the noiseless setting; the online performance of methods is shown in Figure~\ref{fig:hpo_budget}.}
    \label{fig:hpo_cifar}
    \vspace{-.2in}
\end{figure}

However, as we identify in this work, several additional sources of noise may be present during federated evaluation, such as data and systems-related heterogeneity and privacy noise. These noise forms fundamentally alter the evaluation process and, as a result, the performance of federated hyperparameter tuning methods (Figure~\ref{fig:hpo_cifar}).

Although prior work has identified the issue of noisy evaluation~\citep{khodak_federated_2021,wang_fedhpo-b_2022}, the impact and magnitude of this issue on HP tuning remains unclear. In this work, we systematically study the use of noisy evaluation in federated HP tuning. Our study provides insights into best practices for federated HP tuning and suggests several directions for further study in this broad area. Our results also lead us to propose simple baselines that can help to mitigate the effect of noisy evaluation in practical FL applications. Overall, we make the following contributions:

\vspace{-.1in}
\begin{itemize}[leftmargin=*]
\setlength\itemsep{0em}
    \item We identify and systematically explore key sources of noise in federated evaluation, including client subsampling, data, and systems heterogeneity, data privacy, and the use of proxy data. We focus on cross-device FL (e.g., learning across hundreds to millions of phones). However, our insights (particularly around privacy noise and proxy data) may also extend to cross-silo settings (e.g., learning across tens of hospitals).
    
    \item Across a range of large-scale federated learning datasets, we show that even small amounts of noise in the evaluation process can significantly degrade the performance of hyperparameter tuning methods. Our results highlight best practices for practical federated learning applications (e.g., reverting to simple baselines in high-noise settings) and demonstrate a need for future study in this area.
    
    \item Finally, we propose a simple approach for performing hyperparameter tuning in high noise settings based on the use of public proxy data. When available, our results show that hyperparameter tuning on proxy data can be a particularly effective solution in federated networks.
\end{itemize}

\begin{figure*}
    \centering
    \includegraphics[width=8cm]{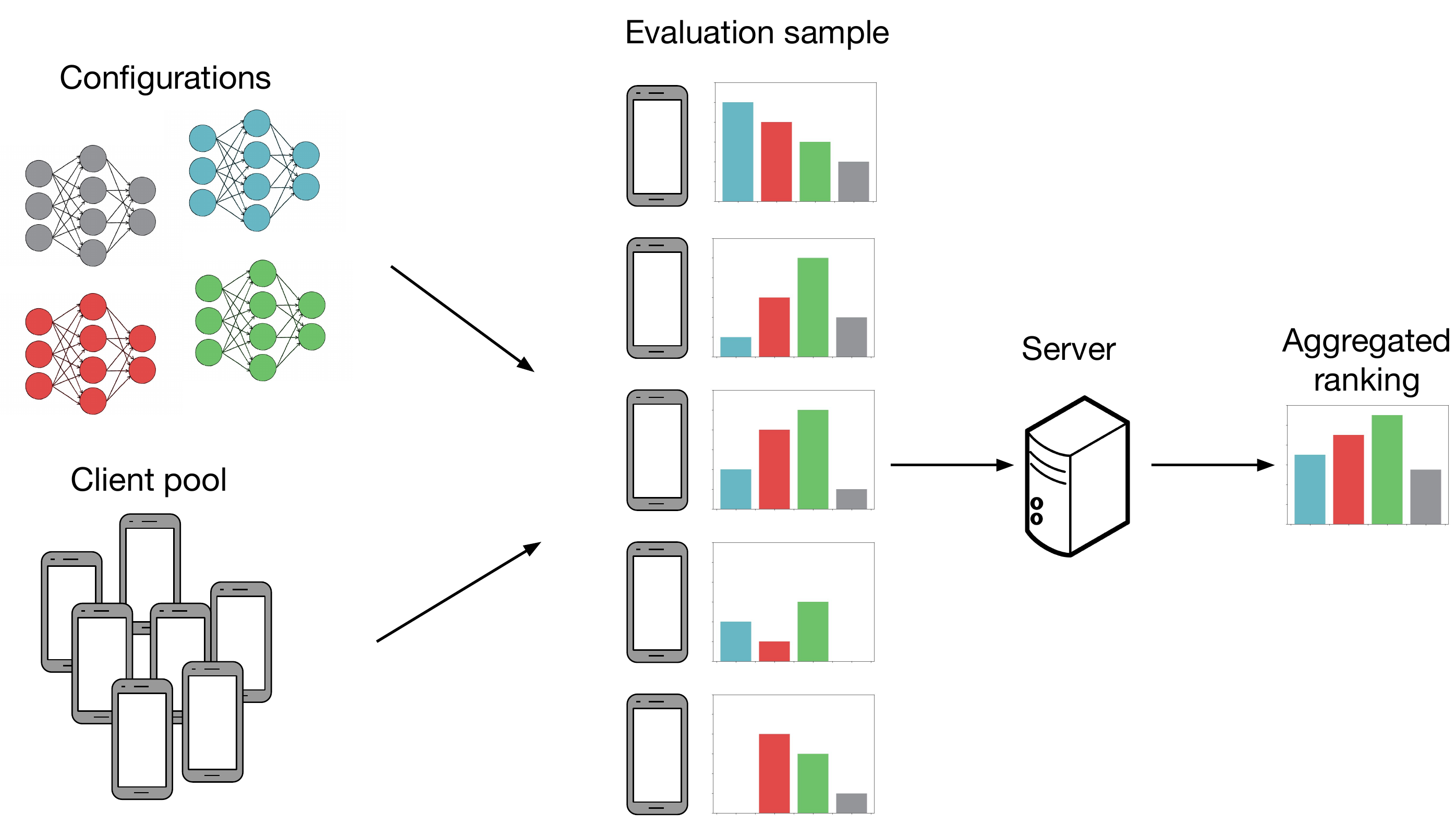}\hspace{0.8cm}
    \includegraphics[width=8cm]{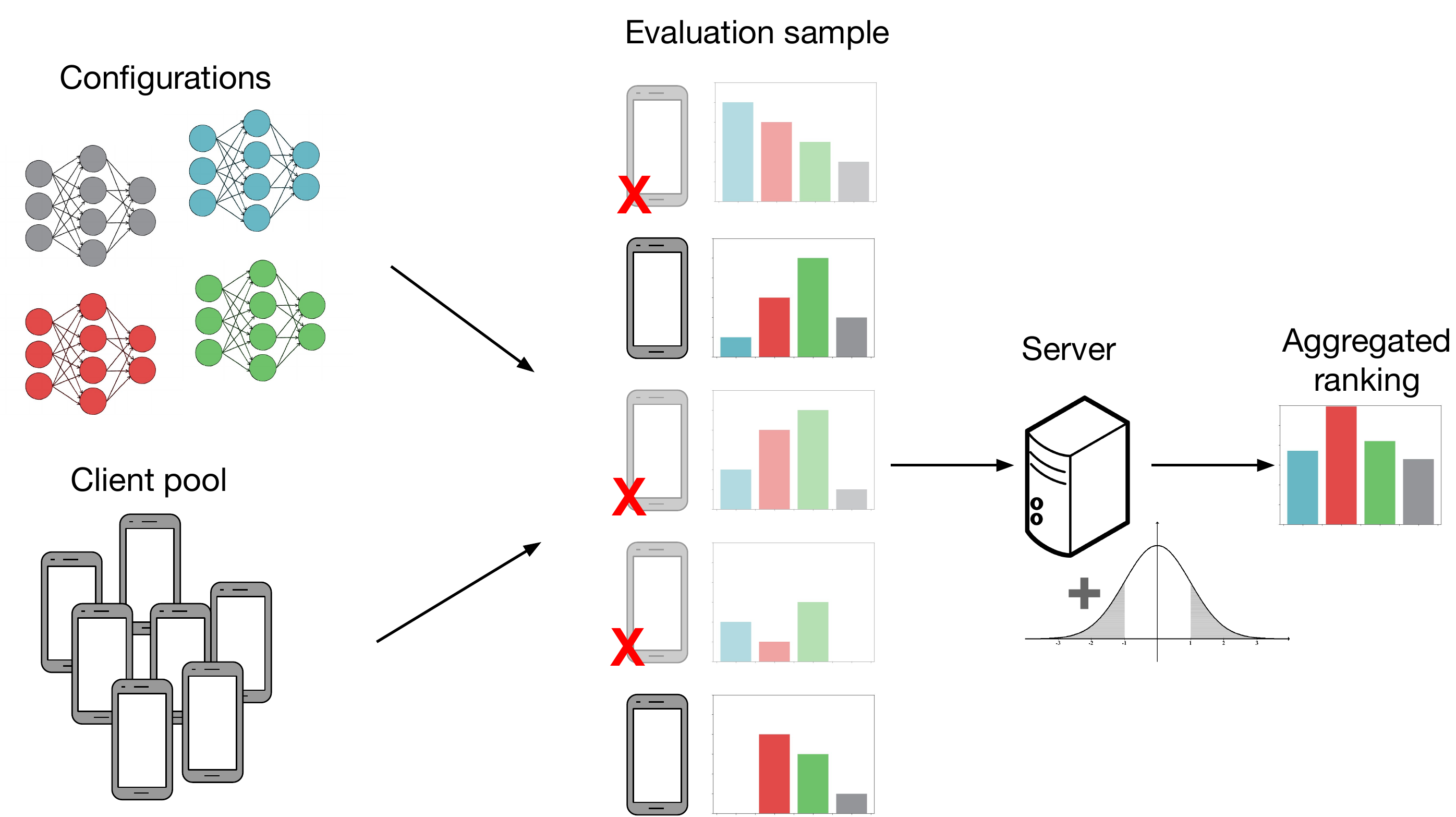}
    \vspace{-.05in}
    \caption{Several factors contribute to noisy evaluation in cross-device federated learning. In the low-noise setting on the left, the green configuration is correctly identified as the best-performing one. In the high-noise setting on the right, the presence of client subsampling, client heterogeneity, and differential privacy results in a noisy evaluation which incorrectly ranks the red configuration over the green one.}
    \label{fig:SETUP}
    \vspace{-.1in}
\end{figure*}

\vspace{-.2in}
\section{Noisy Evaluation in FL}
\label{sec:background}
\vspace{-.05in}
We begin by taking a closer look at the process of federated hyperparameter tuning: We give an overview of the cross-device FL training/evaluation workflow (\S\ref{sec:bg:fl}), 
identify key sources of evaluation noise  (\S\ref{sec:bg:noise}),
and summarize prior approaches for hyperparameter optimization  (\S\ref{sec:bg:hp}). We discuss closely related works throughout this section, and defer a detailed discussion of prior work to Section~\ref{sec:relwork}.

\vspace{-.1in}
\subsection{Cross-Device Federated Learning}
\label{sec:bg:fl}
\vspace{-.05in}
Federated learning (FL) considers collaboratively training a machine learning model across a distributed network of clients. In this work we focus on applications of \textit{cross-device} federated learning, which aim to learn across massive networks of remote clients such as mobile phones or wearable devices. For these applications, avoiding the need to centralize data  can be critical to reduce communication and storage costs as well as improve privacy~\cite{mcmahan_communication-efficient_2017,li_federated_2020,kairouz_advances_2021}.

Although many works have studied how issues in cross-device FL such as client subsampling, heterogeneity, and privacy impact \textit{training}, few have studied their effect on \textit{evaluation} (see Section~\ref{sec:relwork}). This is particularly problematic because state-of-the-art approaches for FL often  rely on evaluating additional hyperparameters in the training process (e.g. for momentum and regularization), despite having relatively strict evaluation budgets~\cite{khodak_federated_2021}. Developing methods that can efficiently and effectively evaluate/select HP configurations in FL is thus a key area of practical importance~\cite{kairouz_advances_2021}.

\vspace{-.1in}
\paragraph{Federated HP Tuning.} 

Due to the scale of the network and potentially small client datasets, the prevailing procedure in cross-device FL is to split the data \textit{by client}~\cite{bonawitz_towards_2019,yuan_what_2022}. In this work, the training and validation datasets $D_{\text{tr}}$ and $D_{\text{val}}$ are partitioned across two disjoint client pools of size $N_{\text{tr}}$ and $N_{\text{val}}$ respectively. Furthermore, practical constraints on communication and client availability limit each training/validation round to sampling (without replacement) a subset of these clients. 

Because clients are randomly sampled, it is highly unlikely that either the training or evaluation dataset is consistent across rounds. More precisely, a typical federated \textit{training} algorithm with hyperparameters $\theta$ optimizes model parameters $w$ to minimize a weighted sum of the training clients' losses (Eq.~\ref{eq:fl_train}). Due to the aforementioned constraints in cross-device FL, in practice the loss is estimated at each round using a subsample ($S \subset [N_\text{tr}]$) of the training population ($S = [N_\text{tr}]$).
\begin{align}
\label{eq:fl_train}
    F_{\text{tr}}(w) = \sum_{k\in S} p_{\text{tr},k} F_{\text{tr},k}(w) \;\big/ \; \sum_{k\in S} p_{\text{tr},k}
\end{align}
During \textit{hyperparameter (HP) tuning}, $\theta$ is tuned to minimize a similar weighted sum
\footnote{We evaluate HPs in two settings: \emph{uniform} ($p_{\text{val},k}=1 \; \forall k$) and \emph{weighted} ($p_{\text{val},k} = $ the number of samples on validation client $k$). For all experiments involving differential privacy, we use the uniform evaluation as to bound evaluation sensitivity independently of any client's local dataset size. Otherwise, we evaluate models with the weighted objective. During training, we set $p_{\text{tr},k}$ to match the same scheme (uniform or weighted) as evaluation.}
of validation clients' errors (Eq~\ref{eq:fl_eval}). The ideal procedure for tuning $\theta$ is to generate a set of candidate configurations, evaluate them on the validation population ($S = [N_\text{val}])$, and then select the best-performing one. However, like in training, practical systems are limited to a subset of the validation clients $S \subset [N_\text{val}]$.
\begin{align}
\label{eq:fl_eval}
    F_{\text{val}}(\theta) = \sum_{k\in S}p_{\text{val},k} F_{\text{val},k}(w(\theta)) \;\big/ \; \sum_{k\in S} p_{\text{val},k}
\end{align}
For the purposes of this work, we assume that $\theta$ is \textit{global} i.e. shared across all clients. Although tuning hyperparameters which are \textit{personalized} to specific clients has been a focus of prior work in FL (see Section~\ref{sec:relwork}), supporting such methods in cross-device FL may be impractical when partitioning the data by client. Here, we instead start with the simpler problem of optimizing global hyperparameters and see that this already seemingly simple procedure can become exceptionally difficult in light of noisy evaluation. 

\vspace{-.1in}
\subsection{Sources of Evaluation Noise in FL}
\label{sec:bg:noise}
Below, we further describe challenges of subsampling and introduce two other sources of noise: heterogeneity and privacy. As summarized in Figure~\ref{fig:SETUP}, these sources of noise can contribute to noisy and unreliable evaluations.

\paragraph{1. Subsampling.} Production cross-device FL systems 
can face strict constraints on communication and client-side computation. Additionally, clients themselves may need to satisfy certain conditions (e.g., the phone being idle, connected to WiFi, and charging, as indicated by phones without red `X's in Figure~\ref{fig:SETUP}) in order to participate in evaluation~\cite{bonawitz_towards_2019}. For these reasons, it is impractical to access all $N_{\text{val}}$ validation clients and obtain a \textit{full evaluation} (Eq.~\ref{eq:fl_eval}, $S = [N_\text{val}]$)
for every HP configuration. Instead, we assume access to a noisy evaluation reported by a \textit{subsampled} set of validation clients $S \subset [N_\text{val}]$.


\vspace{-.1in}
\paragraph{2. Heterogeneity.} As each device in a federated network generates its own local data, \textit{data  heterogeneity} (i.e., non-identically distributed data between clients) is a common concern in FL~\cite{li_federated_2020,kairouz_advances_2021}. Such heterogeneity may occur, for example, due to differing locations, linguistic styles, or usage patterns from one client to another. As we show in Section~\ref{sec:exp:heterogeneity}, heterogeneity in the data can take an already `noisy' evaluation sample and bias it further, as  any two clients may rank the same set of configurations differently. In our experiments, we demonstrate this effect in both natural and synthetic datasets.  

Beyond data heterogeneity, FL networks are also prone to issues of \textit{systems heterogeneity}, which refers to varying participation capabilities across clients due to to differences in hardware, network quality, and device availability \citep{li_federated_2020-1}. These conditions may present as another source of bias dictating how frequently clients participate in evaluation. As a result, hyperparameter tuning algorithms may be naturally biased towards selecting configurations which perform well on high-participating clients, not necessarily on the entire population (see Section~\ref{sec:exp:heterogeneity}).

\vspace{-.1in}
\paragraph{3. Privacy.} Finally, a key concern in FL is the privacy of clients’ data. 
The predominant form of privacy considered in cross-device FL is \textit{client-level differential privacy}~\cite{dwork_algorithmic_2013, brendan2018learning}, which at a high level aims to mask whether or not a given client has participated in training and/or validation. 
It is important to enforce privacy not only at training time~\cite{abadi_deep_2016}
but also in the hyperparameter tuning process, 
as the model selected can itself leak information about the clients that participated in the tuning and validation process ~\cite{papernot_hyperparameter_2022, liu_private_2019, chaudhuri_differentially_2011}.
In order to make the hyperparameter tuning algorithm private,
the server perturbs the aggregate evaluation statistic (e.g., the accuracy) of each configuration with Laplace noise
at each iteration of the tuning procedure.
Similar to issues of heterogeneity, client subsampling affects the noise introduced by privacy. 
When more clients participate in evaluation, the averaged evaluation loss is less sensitive to the loss of any one particular client, and thus requires adding less noise to achieve the same level of privacy. 
As we show in Section~\ref{sec:exp:privacy}, even generous privacy budgets can make  evaluation  extremely noisy. In Section~\ref{sec:proxy} we explore the use of public proxy data (which may also be considered its own, separate form of `noise') to address high-noise evaluation. 

\begin{figure*}[t!]
    \centering
    \includegraphics[width=17cm]{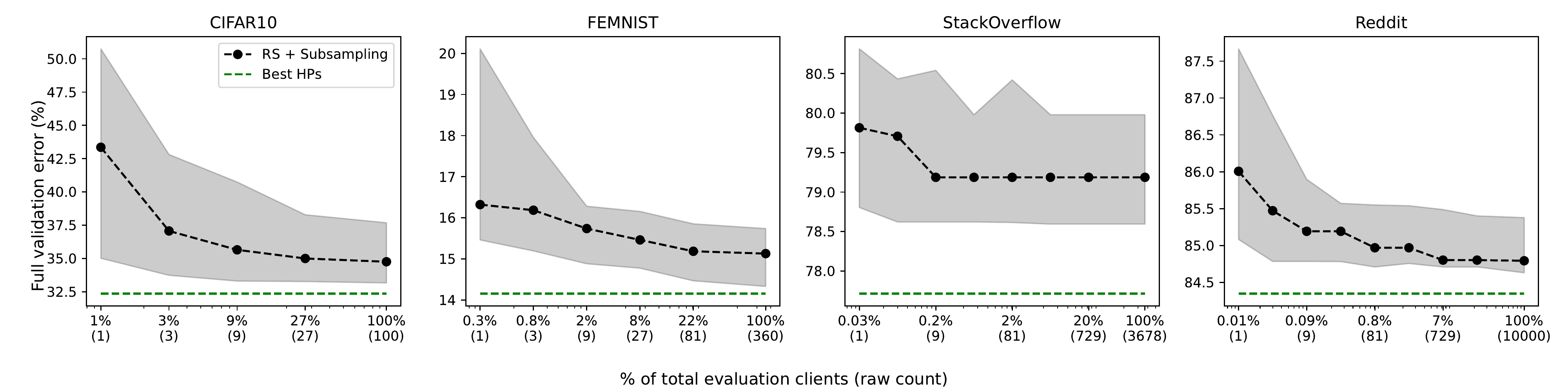}
    \vspace{-.05in}
    \caption{We run random search (RS) with a fixed budget $(K=16 \text{ configurations})$ while varying the \textit{subsampling rate} from a single client to the full validation client pool. We evaluate the error rate of the configuration found by RS on all clients and report a \textit{weighted} average. We plot the median and fill in quartiles over RS trials. ``Best HPs'' shows performance of the best RS trial at full evaluation. Larger datasets can mitigate the problem of subsampling as we can sample a larger raw number of clients.}
    \label{fig:subsampling}
    \vspace{-.1in}
\end{figure*}

\vspace{-.1in}
\subsection{Hyperparameter Tuning Methods}
\label{sec:bg:hp}
\vspace{-.05in}
Given a HP search space and overall budget, HP tuning methods aim to find configurations in the search space that optimize some measure of quality (e.g., minimize error rate) within a constrained budget (e.g., computational cost). 

Classical HP tuning methods generate candidate HP configurations over a grid (grid search) or at random (random search). Each configuration is used to perform some predetermined training routine, e.g., training for a fixed number of epochs or until some fixed stopping criterion is achieved~\cite{bergstra_random_2012}. Subsequently, each configuration is evaluated and the best performing one is returned. There exist two main strategies for improving upon these classical approaches: adaptively \textit{generating} configurations  (e.g., Bayesian optimization approaches), or  adaptively \textit{evaluating} configurations (e.g., early stopping approaches). 
In this work we explore representative candidates from each category of HP tuning method---using random search (RS) as a classical/simple baseline which we compare to a more sophisticated Bayesian optimization (TPE), early stopping (Hyperband), and hybrid approach (BOHB)~\cite{bergstra_algorithms_2011,li_hyperband_2017,falkner_bohb_2018}.  Prior work in federated HP tuning often uses these classes HP tuning methods, but they do not explore the effect of noisy evaluation, which we discuss further in Section~\ref{sec:relwork}. We provide a detailed description of these methods in Appendix~\ref{sec:appendix:hpo}, as well as pseudocode for RS (Algorithm~\ref{alg:rs}) and its FL counterpart (Algorithm~\ref{alg:rs_fl}) in Appendix~\ref{sec:appendix:pseudocode}.
\vspace{-.1in}
\section{Experiments}
\label{sec:exps}
\vspace{-.05in}
In this section we present experiments detailing the effect of noisy evaluation in federated settings. By analyzing these sources of noise individually and in combination with one another, we aim to answer the following questions:

\textbf{Question 1:} \textit{To what extent does subsampling validation clients degrade the performance of HP tuning algorithms?}

\textbf{Question 2:} \textit{How, and to what extent, do the factors of data heterogeneity, systems heterogeneity, and privacy exacerbate  issues of subsampling?}

\textbf{Question 3:} \textit{In noisy settings, how do popular HP tuning algorithms compare to simple baselines?}

\vspace{-.1in}
\paragraph{Datasets.} We optimize HPs of deep learning models on several standard FL benchmarks: CIFAR10 \citep{krizhevsky_learning_2009}, FEMNIST \citep{caldas_leaf_2018}, StackOverflow \citep{authors_tensorflow_2019}, and Reddit\footnote{We use the December 2017 Reddit data from a larger pre-existing dataset publicly available from pushshift.io.}~\citep{caldas_leaf_2018}. 
We follow the method in \citet{hsu_measuring_2019} of synthetically partitioning CIFAR10 according to a Dirichlet distribution with parameter $\alpha=0.1$ in order to generate imbalanced client labels. The other three datasets have natural client partitions. We provide summary statistics in Table~\ref{tab:datasets}, while Table~\ref{tab:appendix:datasets} in the appendix contains more detailed information.

\begin{table}[h!]
    \centering
    \begin{tabular}{lcccc}
        \toprule[\heavyrulewidth]
        & \multicolumn{2}{c}{\#Clients} & \multicolumn{2}{c}{\#Examples} \\
        \textbf{Dataset} & \textbf{Train} & \textbf{Eval} & \textbf{Mean} & \textbf{Total} \\
        \midrule
        CIFAR10 & 400 & 100 & 100 & 5K \\
        FEMNIST & 3.5K & 360 & 203 & 73K \\
        StackOverflow & 10.8K & 3.7K & 391 & 5.6M\\
        Reddit  & 40K & 10K & 19 & 1.1M\\
         \bottomrule[\heavyrulewidth]
    \end{tabular}
    \caption{Statistics of the datasets used in the experiments.}
    \label{tab:datasets}
    \vspace{-.2in}
\end{table}

\paragraph{Training.} On CIFAR10 and FEMNIST, we train 2-layer CNNs to perform image classification. For StackOverflow and Reddit we tokenize the text using the GPT2 tokenizer \citep{radford_language_2019} and train a 2-layer LSTM with an embedding and hidden size of 128 to predict the next token in a sequence with a maximum length of 25 tokens. On all datasets, we uniformly sample 10 clients per training round. We use FedAdam~\cite{reddi_adaptive_2020} as the FL optimizer and keep its HPs fixed within individual training runs.

\vspace{-.1in}
\paragraph{Hyperparameters.} We tune five HPs (search space in Appendix~\ref{sec:appendix:hpspace}): three server FedAdam HPs (learning rate, 1\textsuperscript{st} and 2\textsuperscript{nd} moment decay rates) and two client SGD HPs (learning rate and batch size). These are a natural set of HPs to explore in the context of FL, with the client/server learning rate and batch size being present in virtually all federated optimization methods~\cite{wang_field_2021}, and the Adam-specific HPs having been shown to yield significant improvements in practice~\cite{reddi_adaptive_2020}. 

As discussed in Section~\ref{sec:bg:fl}, client HPs are not personalized, i.e., all clients share the same learning rate and batch size. As mentioned in Section~\ref{sec:bg:hp}, we evaluate a representative set of methods: random search (as a simple baseline), Hyperband (an early stopping method), Tree Parzen Estimator (a Bayesian optimization method), and BOHB (a hybrid of TPE and HB). Each method is allocated a total budget of $6480$ training rounds and a maximum of $405$ rounds per HP configuration. RS and TPE search $K=16$ configurations, while Hyperband and BOHB search through 5 brackets of SHA with an elimination factor $\eta=3$. 

\begin{figure*}[t!]
    \centering
    \includegraphics[width=17cm]{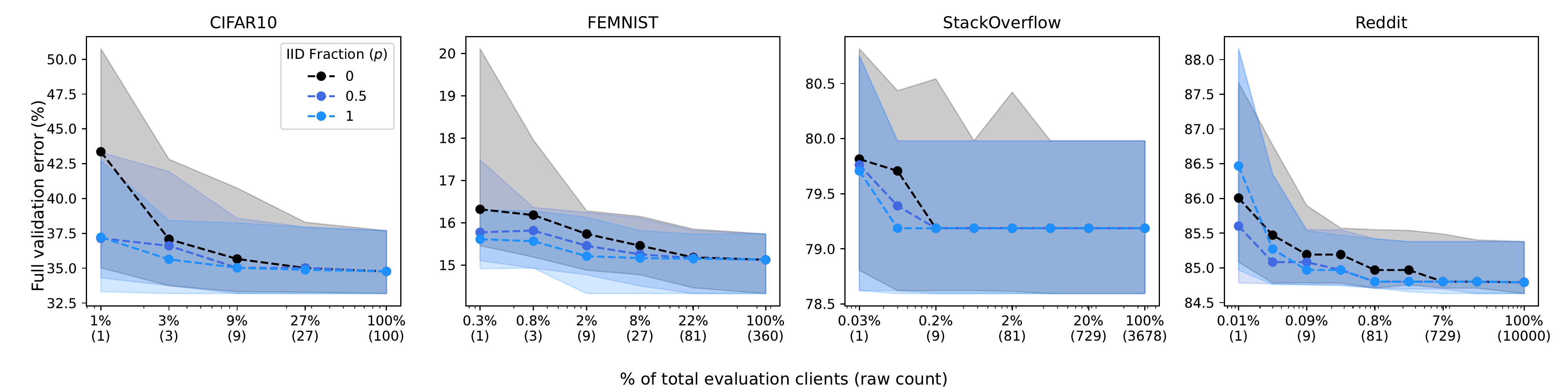}
    \caption{We run RS on three separate validation partitions with varying degrees of \textit{data heterogeneity}. Client subsampling generally harms performance, but has a greater impact on performance when the data is heterogeneous ($p=0$) rather than homogeneous $(p=1.0)$.}
    \label{fig:niid}
    \vspace{-.1in}
\end{figure*}

\begin{figure}[h!]
    \centering
    \includegraphics[width=8cm]{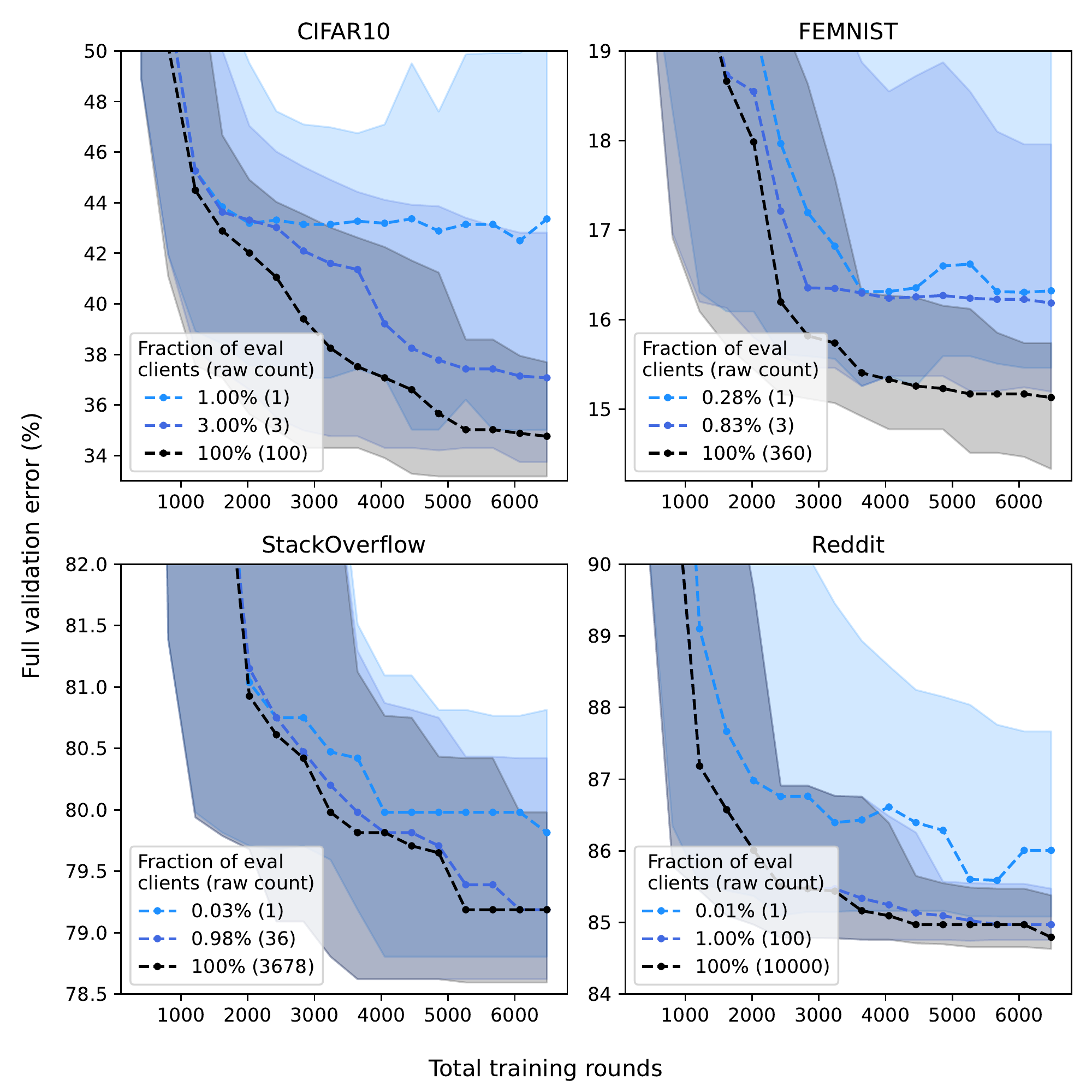}
    \caption{Performance of RS as the training budget is used up. Subsampling evaluation clients harms the performance-budget tradeoff, and the gap between subsampling and full evaluation grows as more of the budget is allocated.}
    \label{fig:sub_budget}
    \vspace{-.2in}
\end{figure}

\paragraph{Evaluation.} In RS-only figures, we train random 128 HP configs and then bootstrap 100 trials i.e. run RS on $K=16$ HP configs that are resampled from the set of 128. In all plots, we show the median \textit{full validation \% error} (Eq.~\ref{eq:fl_eval}, $S = [N_\text{val}])$ and fill in the lower/upper quartiles. In other words, all $N_{\text{val}}$ validation clients are used to perform ``testing'', thus reusing the subset of clients sampled during HP evaluation. We chose this form of evaluation as practical cross-device settings may lack client partitions~\cite{bonawitz_towards_2019}.

\subsection{Client subsampling}
\label{sec:exp:subsampling}
\paragraph{Observation 1:} \textit{High degrees of subsampling hurt HP tuning performance.}

To explore the effect of subsampling, we run random search across all datasets while varying the evaluation client sampling rate. Figure~\ref{fig:subsampling} shows that subsampling increases the median error rate by up to 8\% on CIFAR10 and up to 2\% on the other datasets. The upper quartile values increase even more (e.g. 12\% on CIFAR10), showing less reliable performance. To recover performance levels close to full evaluation on these datasets, sampling $\sim$100 clients is sufficient, which is a favorable sign for production settings that assume a small percentage but sizable raw number of clients are available during a given round.

\vspace{-.2in}
\paragraph{Observation 2:} \textit{Allocating additional training budget can mitigate the effects of subsampling, but only to an extent.} 

Due to resource limitations in federated learning, we are concerned with not only the quality, but also the cost of finding the best configuration. To show the tradeoff between these two variables, we record the performance of RS as the search budget (in training rounds) is used to train 16 configurations. Figure~\ref{fig:sub_budget} shows these  curves when RS evaluations are performed with different subsampling rates. While all runs start with similar performance, the gap between subsampling and full evaluation grows as more of the budget is allocated, eventually leading  to the final performance gaps previously shown in Figure~\ref{fig:subsampling}. We observe that client subsampling harms not only the final performance of RS, but also its overall accuracy-budget tradeoff. On all four datasets, sampling a single client harms convergence.

\begin{figure*}[t!]
    \centering
    \includegraphics[width=17cm]{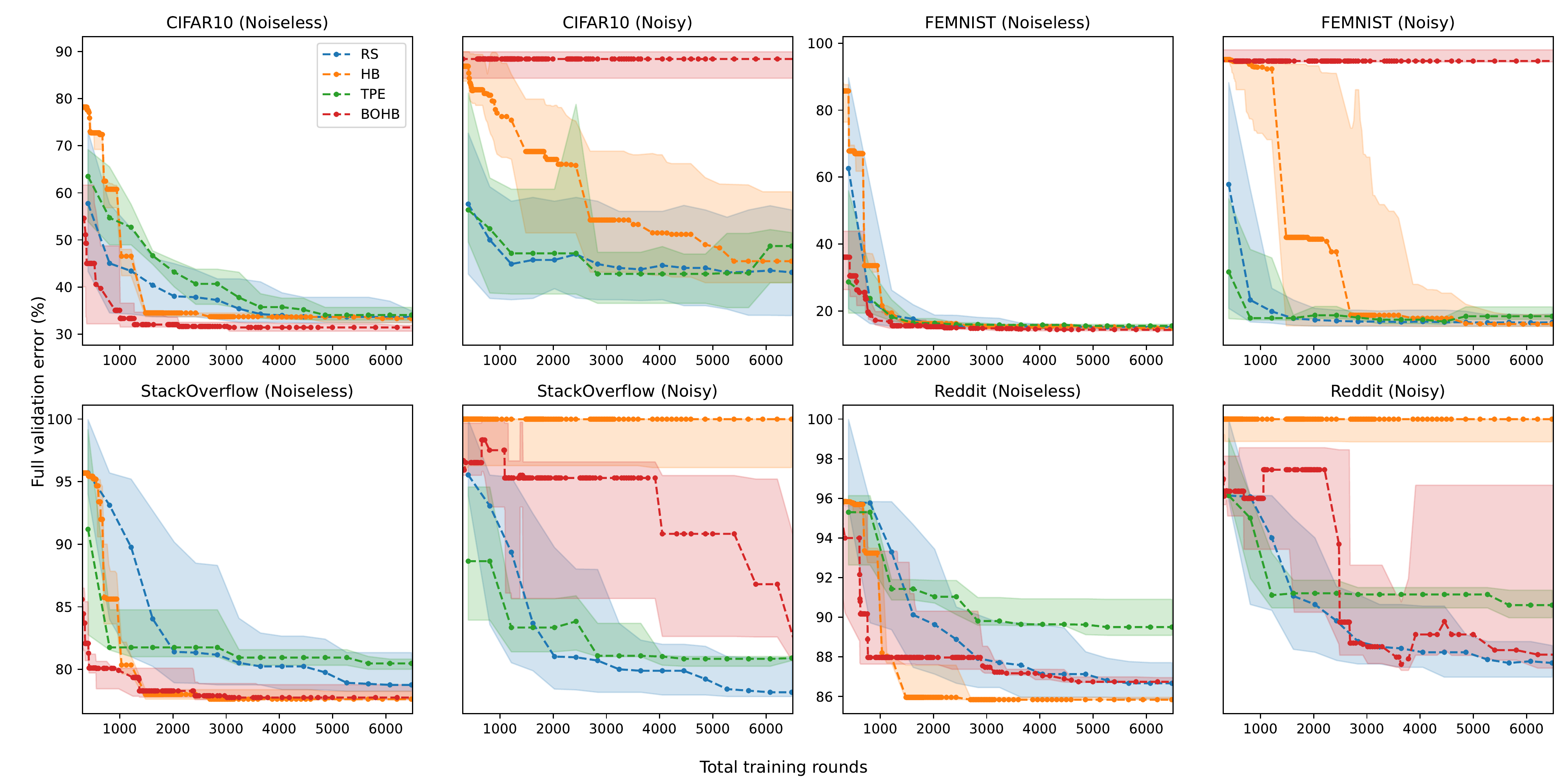}
     \vspace{-.1in}
    \caption{Performance of RS, HB, TPE, and BOHB in noiseless versus noisy (subsampling 1\% of clients, $\eps=100$ privacy) settings. In several cases, the quality of HB or BOHB degrades to that of random guessing. We plot the median and quartiles over 8 trials.}
    \label{fig:hpo_budget}
    \vspace{-.1in}
\end{figure*}

We note that there are several ways to measure the HP tuning budget, such as the number of train/eval rounds \citep{khodak_federated_2021}, wall-clock time \citep{li_hyperband_2017}, or computation load \citep{zhang_fedtune_2022}. For simplicity, we do not consider time spent on evaluation rounds and server-side optimization. We note that TPE, HB, and BOHB use more of these two resources compared to RS. Despite our evaluation being advantageous to these methods, we find they underperform against RS at higher levels of noise (see Figure~\ref{fig:hpo_budget}). Still, it is important to investigate efficient HP tuning under different types of resource budgets in FL, a discussion of which we defer to Section~\ref{sec:discussion}.

\begin{figure*}[t!]
    \centering
    \includegraphics[width=17cm]{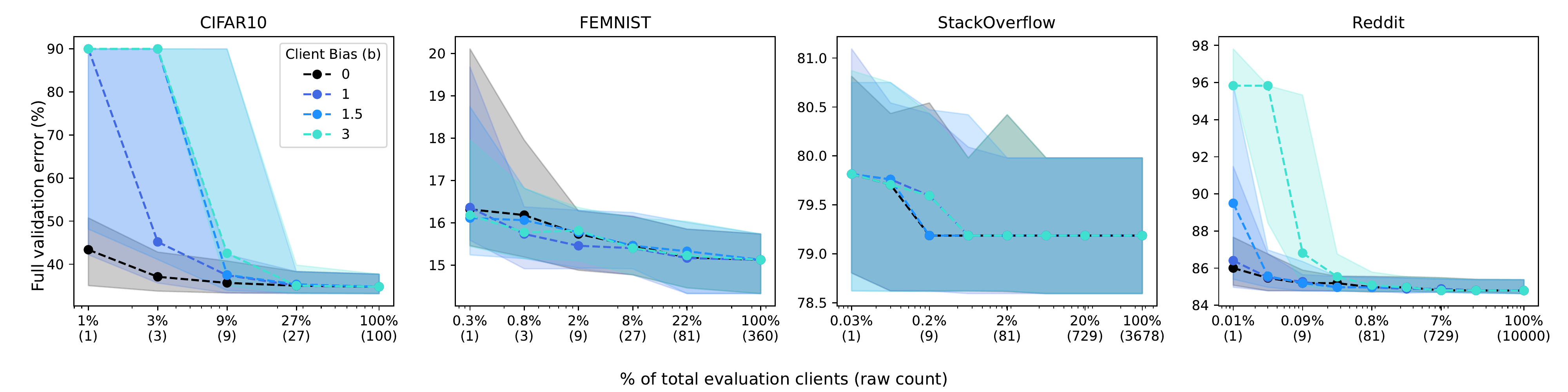}
    \caption{We run RS on each dataset and bias the client sampling to reflect four degrees of \textit{systems heterogeneity}. On CIFAR10 and Reddit, performance degrades as sampling becomes more biased towards better-performing clients.}
    \label{fig:SYS}
    \vspace{-.1in}
\end{figure*}

\begin{figure}[h!]
    \centering
    \vspace{-.15in}
    \includegraphics[width=7cm]{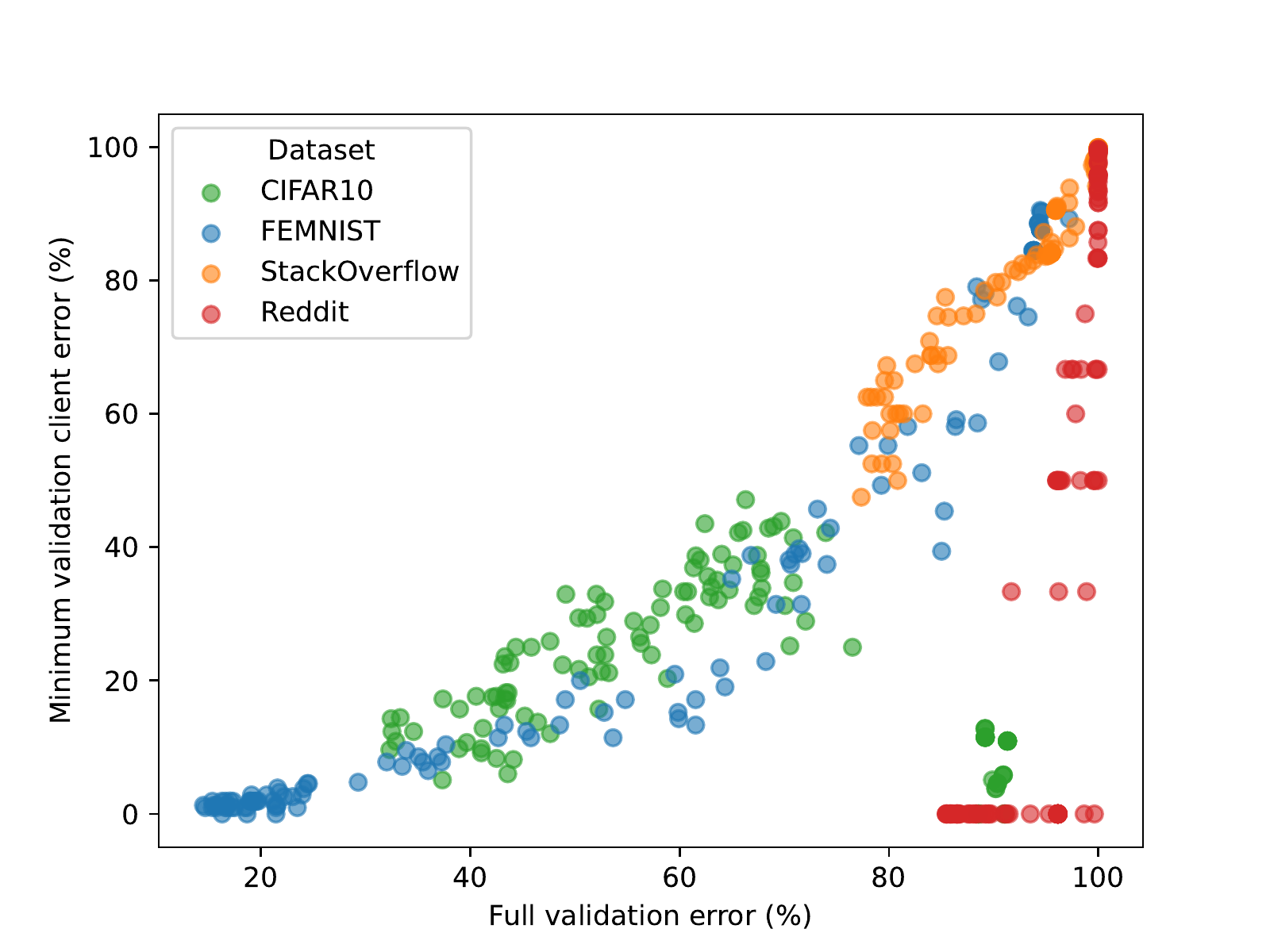} 
    \caption{We plot each configuration at the coordinates ($x=\text{full evaluation error}, y=\text{minimum client error}$). On CIFAR10 and Reddit, many configurations have poor global performance but extremely good performance on a few clients.}
   \vspace{-.1in}
    \label{fig:sys_scatter}
\end{figure}

\subsection{Heterogeneity}
\label{sec:exp:heterogeneity}
\paragraph{Observation 3:} \textit{Data heterogeneity exacerbates the negative effects of subsampling.} 
\vspace{-.1in}
\paragraph{Data heterogeneity.} We analyze multiple forms of data heterogeneity by running experiments on datasets with both synthetic (CIFAR10) and natural (FEMNIST, StackOverflow, Reddit) partitions. In addition to testing multiple datasets, we aim to quantify the impact of heterogeneity by comparing \textit{iid} and \textit{non-iid} versions of the same dataset. We keep the training data in its original partition and repartition the evaluation client data. To repartition a naturally heterogeneous \textit{(non-iid)} federated dataset into a homogeneous \textit{(iid)} version, we pool all of the eval data and let each eval client resample the data in an iid manner. More specifically, all clients share a distribution where each data point of the pooled dataset is equally likely to be sampled \citep{caldas_leaf_2018}. We extend this method by resampling only a fraction $p \in [0, 1]$ of the validation data, which allows us to vary the level of heterogeneity from naturally non-iid $(p=0)$ to artificially iid $(p=1)$. We design our data heterogeneity experiments on a single dataset: The evaluation client data is repartitioned at three levels of data heterogeneity $p \in \{0, 0.5, 1\}$. We run RS at multiple subsampling rates on each of the three partitions.

We present the results in Figure~\ref{fig:niid}. First, varying heterogeneity has no effect in the full evaluation setting. Second, across all subsampling rates, RS, on average, finds better configurations when running evaluations on the iid partition compared to the non-iid partition. Finally, noisy evaluation degrades performance even when subsampling on the $p=1$ partition. We expect this degradation as a single client does not capture the signal of the entire validation population. 

\paragraph{Observation 4:} \textit{Systems heterogeneity can be catastrophic when there is sufficient underlying client heterogeneity.}

\paragraph{Systems heterogeneity.}  In practical FL settings, high-end devices may participate in training more often, which can bias model performance towards these devices~\cite {bonawitz_towards_2019}. The same participation bias exists during validation, leading to overly optimistic model evaluations. 

We simulate systems heterogeneity conditions by biasing sampling towards clients who perform well on the current model being evaluated. This bias assigns a weight $(a + \delta)^b$ to each client (normalized to a probability vector), where $a$ is the client's accuracy, $\delta$ is a small constant to ensure non-zero probability, and $b$ controls the degree of sampling bias. We set $\delta=10^{-4}$ and test $b \in \{0, 1, 1.5, 3\}$. Like the prior experiments, we do not modify the training data and only assume biased selection during evaluation.

\begin{figure}[h!]
    \centering
    \includegraphics[width=8cm]{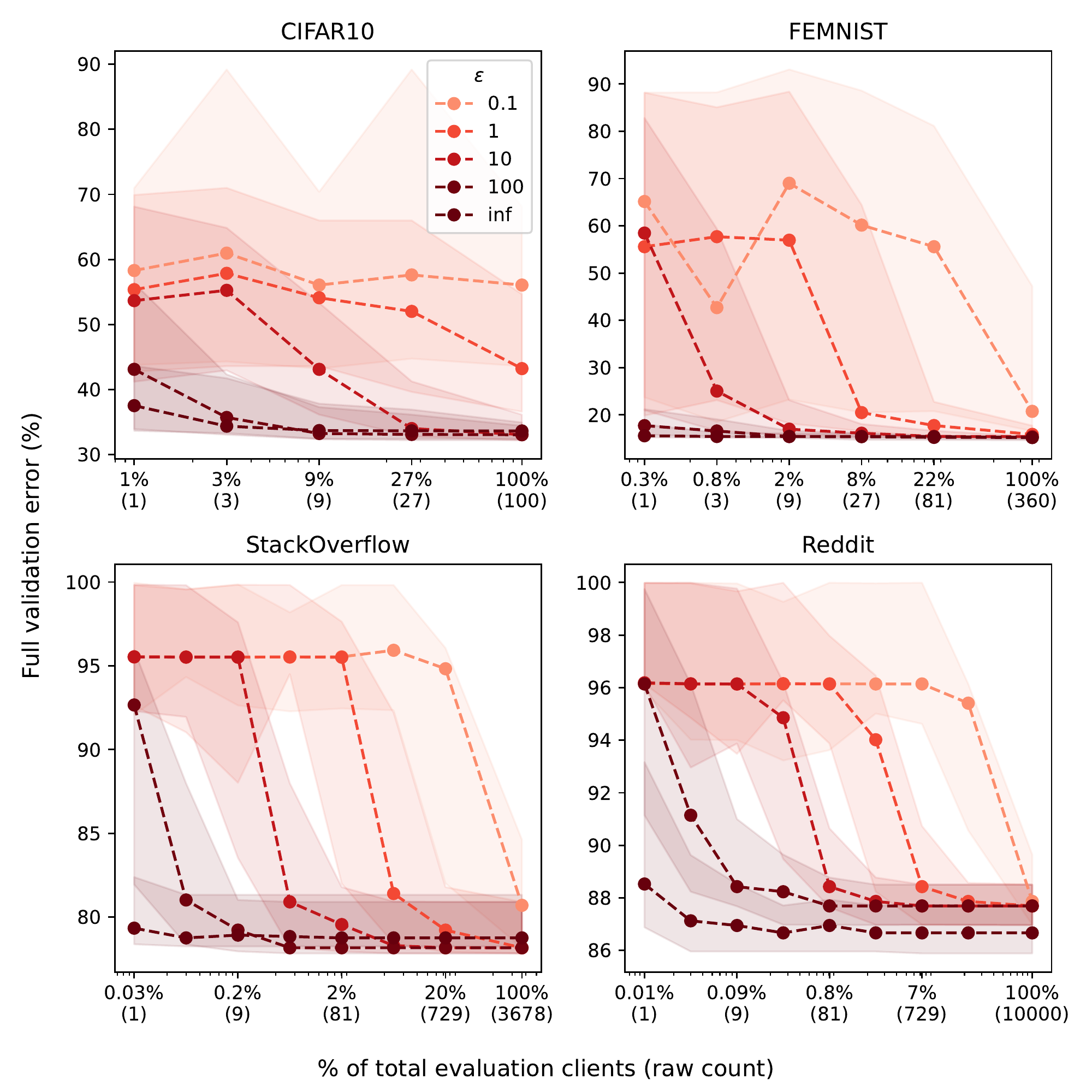}
    \vspace{-.1in}
    \caption{We run RS with 5 different \textit{evaluation privacy} budgets $\eps$. A smaller privacy budget requires sampling a larger raw number of clients to achieve reasonable performance.}
    \label{fig:DP}
    \vspace{-.1in}
\end{figure}

Figure~\ref{fig:SYS} shows the effect of systems heterogeneity combined with lower subsampling rates. Although effects are only noticeable on CIFAR10 and FEMNIST, the drop in performance is catastrophic at low subsampling rates (90\% error rate on CIFAR10). 

We surmise that differences across datasets are due to variations in data heterogeneity. Figure~\ref{fig:sys_scatter} plots 128 configurations with $(x,y)$ coordinates equal to the configuration's (global error, minimum client error) across validation clients. For FEMNIST and StackOverflow, evaluations are `well-behaved' because the variation among clients gradually decreases as the configuration's global performance improves. On the other hand, several configurations trained on CIFAR10 and Reddit have clients with zero error but perform poorly overall, so biasing evaluation towards these clients can be detrimental. These configurations appear in the lower right corner of the plot.

\subsection{Privacy}
\label{sec:exp:privacy}
\paragraph{Observation 5:} \textit{DP noise, even under a generous privacy budget, severely deteriorates performance unless a sufficient number of clients are sampled.} 

In order to understand the impact of differential privacy on HP tuning,
we modify the non-private HP algorithms considered to be differentially private
with respect to the client evaluations.
In particular, each HP tuning method considered operates by evaluating the average accuracy (between 0 and 1) of a
configuration on a set of clients. 
The sensitivity of each evaluation (the impact of a single client on the average accuracy of one configuration) 
is therefore $1/|S|$, where $|S|$ is the size of the set of clients
sampled in each evaluation call.
Preserving privacy for a real-valued query of sensitivity $\Delta$
requires adding Laplace noise with scale $\Delta/\eps$.
The basic composition theorem \cite{dwork_algorithmic_2013}
allows us to allocate a privacy budget of $M/\eps$ to each evaluation,
where $M$ is the total number of evaluations performed.
To satisfy $\eps$-differential privacy, we thus add noise sampled from $\mathrm{Lap}(M/(\eps |S|))$ to each evaluated accuracy.\footnote{Although more sophisticated algorithms for private HP tuning may reduce the overall impact of enforcing privacy,  our goal is to provide a straightforward baseline that demonstrates potential issues that can arise in this setting,
and we therefore implement the simplest mechanism providing pure-$\eps$ DP.}

{
For an algorithm with a total number $T$ of evaluation rounds,
we allocate a privacy budget of $T/\eps$ to each evaluation round. We use the
one-shot Laplace mechanism for top-$k$ selection \cite{qiao_oneshot_2021}
to select the top configurations at each evaluation round.
The one-shot Laplace mechanism adds Laplace noise with scale $\frac{2T k_t}{\eps |S|}$
to the evaluation accuracy for each configuration and releases the identities of the top $k_t$ evaluations at evaluation round $t$.
}

Figure~\ref{fig:DP} shows results of RS when varying the privacy budget and subsampling rate. Noise from privacy clearly hurts performance and makes HP tuning much more challenging than in the non-private setting ($\eps=\text{inf}$). For instance, when subsampling $< 1\%$ of clients on any of the four datasets, applying ($\eps=1$) privacy results in performance similar to randomly choosing HPs. When the privacy level is even more strict ($\eps=0.1$), RS often fails to find good HPs on CIFAR10 even when using all 100 clients for evaluation. On the other datasets, evaluations require least a staggering $30\%$ of clients to avoid this catastrophic degradation in performance.

\vspace{-.1in}
\paragraph{Observation 6:} \textit{In high-noise regimes, popular methods may perform as poorly as naive baselines.}

Finally, we test four HP tuning methods (RS, HB, TPE, and BOHB) in a noisy setting with client subsampling (1\% of population) and DP evaluation ($\eps=100$). Comparing noiseless to noisy evaluation in Figure~\ref{fig:hpo_budget}, we generally see an increase an error rate across all datasets and methods. Furthermore, HB and BOHB disproportionately suffer from subsampling and privacy noise due to the high number of low-fidelity evaluations they use. On each dataset, the \textit{best} method under \textit{noiseless} evaluation (either HB or BOHB) becomes the \textit{worst} under noisy evaluation. 
\begin{figure}[t!]
    \centering
    \includegraphics[width=8cm]{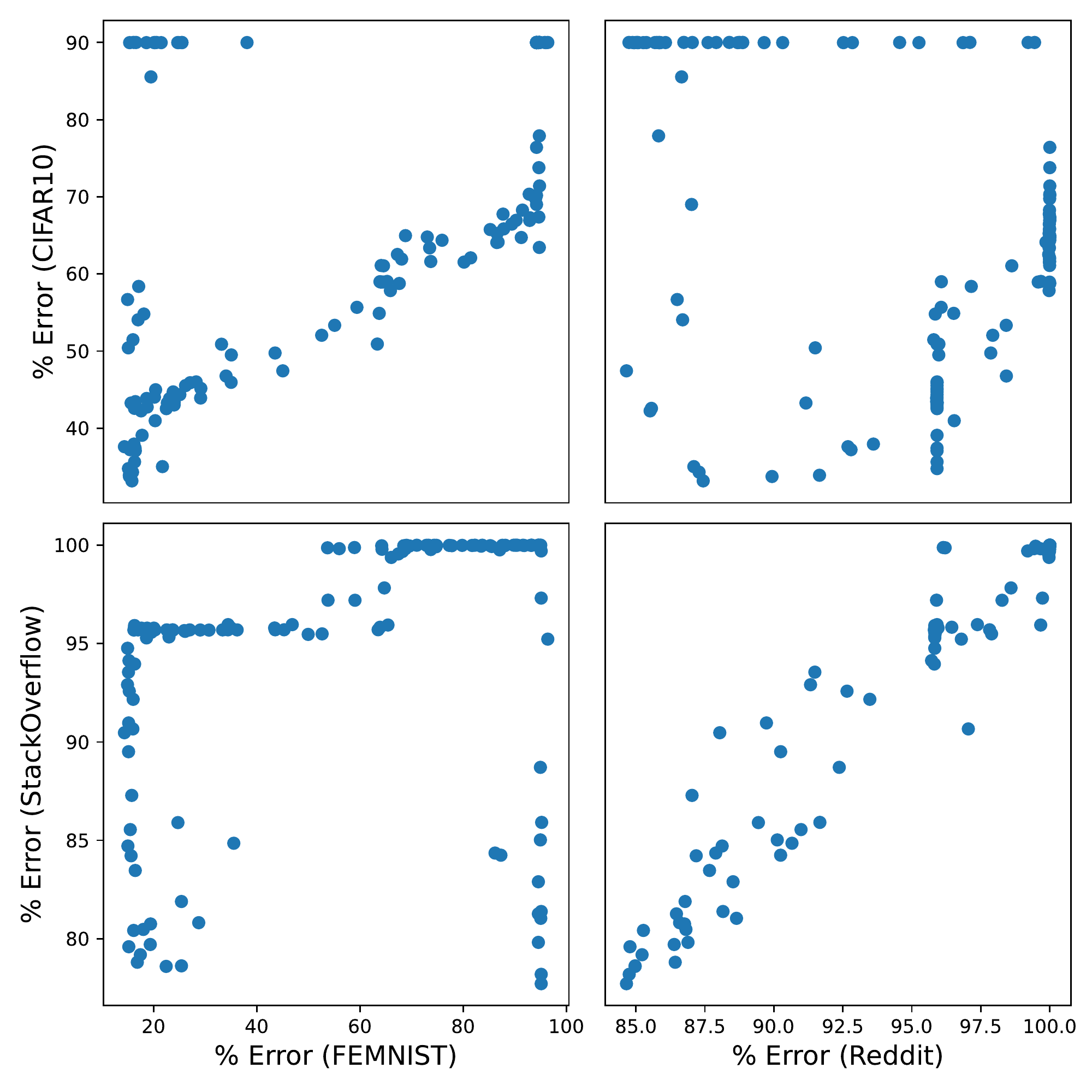}
    \vspace{-.1in}
    \caption{We plot full validation error on 4 pairs of datasets (one pair for each plot). Each of the 128 points represents a single hyperparameter configuration and its $(x,y)$ coordinates show the error of two models separately trained+evaluated on each dataset.}
    \label{fig:proxy_scatter}
    \vspace{-.3in}
\end{figure}

\begin{figure}[t!]
\centering
    \includegraphics[width=8cm]{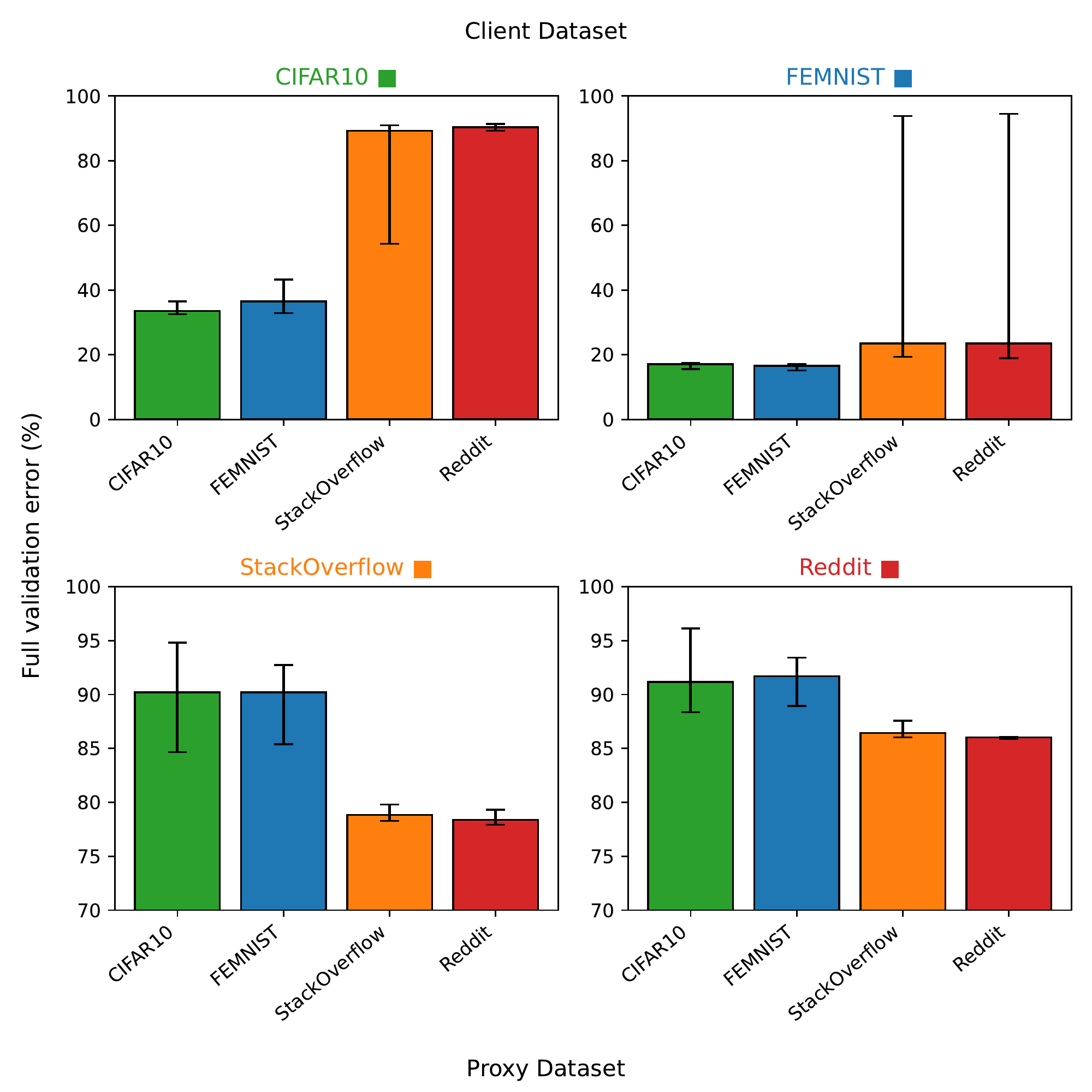}
     \vspace{-.1in}
    \caption{Results of one-shot proxy RS (searching over $K=16$ HP configs using only proxy data) for different pairs of datasets.}
    \label{fig:10PXBAR}
    \vspace{-.1in}
\end{figure}

\vspace{-.1in}
\section{Proxy data}
\label{sec:proxy}
\vspace{-.05in}
In FL settings with extreme resource constraints, server-side \textit{proxy data} can be a valuable source of validation signal, as it allows us to select HPs without accessing \textit{client data}. However, if we rely entirely on proxy data, HP quality is largely determined by the similarity between the proxy and client tasks. Furthermore, it is often difficult to find a suitable proxy dataset for a specific FL dataset. Therefore, we begin by exploring how well HPs transfer across the four datasets used in our experiments.

\begin{figure*}
    \centering
    \includegraphics[width=17cm]{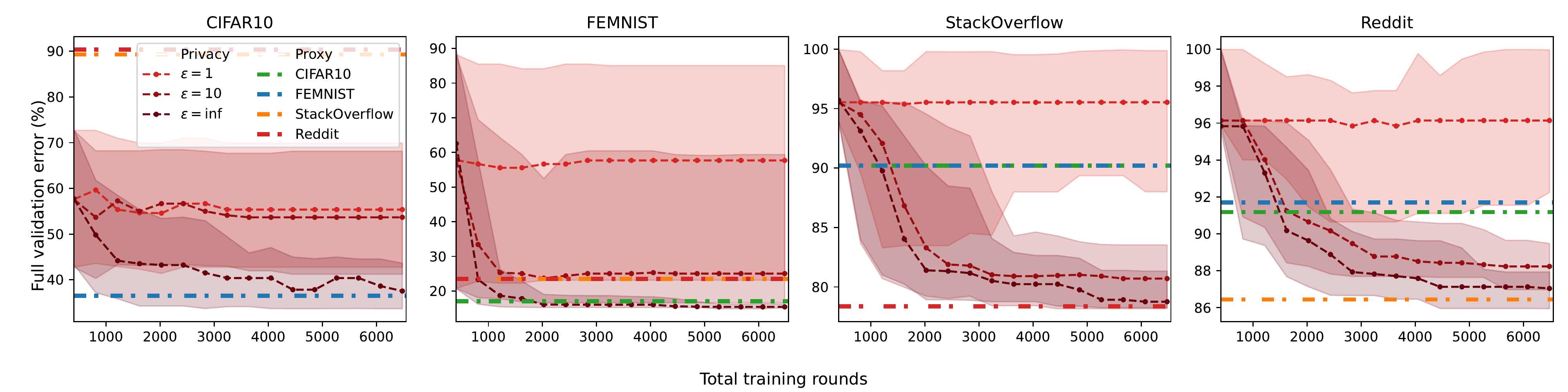}
    \vspace{-.1in}
    \caption{We show the performance vs. budget of RS at multiple values of privacy (subsampling $1\%$ of the validation clients) and compare it to choosing HPs with various proxy datasets (training a single model with the chosen HPs). As evaluation noise increases, datasets which would normally be considered a poor match (e.g. using StackOverflow as a proxy for FEMNIST) become favorable alternatives to using noisy evaluations on the true data.}
    \label{fig:px_budget}
    \vspace{-.1in}
\end{figure*}

\vspace{-.1in}
\paragraph{Observation 7:} \textit{Relying on proxy data can itself be considered a source of noise when there is significant mismatch between proxy and client datasets.}

We consider 4 dataset pairs in Figure~\ref{fig:proxy_scatter}. For a given FedAdam HP configuration, we separately train and evaluate a model on the two datasets. On the (CIFAR10, FEMNIST) and (StackOverflow, Reddit) pairs, HPs can transfer very well. An intuitive reason for this transfer is that these pairs share the same type of task (image classification or next-token prediction) and model architecture (2-layer CNN or LSTM). In light of this observation, we propose a strong two-step baseline which we call \textit{one-shot proxy RS}: 
\vspace{-.1in}
\begin{enumerate}
    \item Run RS using the proxy data to both train and evaluate HPs. We assume the proxy data is both \textit{public and server-side}, so we can always evaluate HPs without subsampling clients or adding DP noise. 
    \item The best configuration found is then used to train a model on the client data. Since we pass only a single configuration to this step, performance is \textit{unaffected} by any sources of evaluation noise in the client data.
\end{enumerate}
\vspace{-.15in}
Surprisingly, the results in Figure \ref{fig:10PXBAR} show that tuning HPs on proxy data is competitive with using the client dataset (even without noisy evaluation). However, as expected, if the datasets are mismatched, performance can become worse than randomly selecting HPs.

\vspace{-.1in}
\paragraph{Observation 8:} \textit{In high-noise regimes, a suitable proxy dataset can assist hyperparameter search.}

In Figure~\ref{fig:px_budget}, we compare HP tuning using noisy evaluations against the one-shot proxy RS method described above. For noisy evaluation, we run RS with a $1\%$ client subsample and vary the evaluation privacy budget. For all datasets, the best possible proxy dataset is competitive with non-private evaluation ($\eps=\text{inf}$).
However, as proxy data is unaffected by noisy evaluation, even a suboptimal proxy dataset can be helpful when evaluation is sufficiently noisy ($\eps=1$).
\vspace{-.1in}
\section{Related Work}
\label{sec:relwork}
\vspace{-.05in}
\paragraph{Federated hyperparameter tuning.}
Prior work in {cross-device} FL identifies resource limitations as a major challenge in HP tuning \citep{kairouz_advances_2021}. Proposed improvements include extending adaptive optimization methods to FL \citep{koskela_learning_2018, reddi_adaptive_2020}, interleaving HP and weight updates during training \citep{mostafa_robust_2019,mlodozeniechyperparameter}, and selecting personalized hyperparameters \citep{agrawal_genetic_2021} for different clients. In addition to resource limitations, other works address data heterogeneity \citep{khodak_federated_2021}, systems heterogeneity \citep{zhang_fedtune_2022}, and privacy \citep{chen_dap_2023} from the perspective of federated HP tuning. 


Existing works also attempt to benchmark HP tuning algorithms on FL datasets. In addition to evaluating a large number of HP tuning methods and datasets, \citet{wang_fedhpo-b_2022} experimentally show that lower sampling rates can mitigate straggler issues in settings with poor network quality. \citet{holly_evaluation_2021} benchmark random/grid search and GP-UCB in an federated learning setting where sufficiently similar clients can share their data with each other.

Finally, another line of work more suited to the \textit{cross-silo} setting has each client perform local hyperparameter tuning and shares their results with other clients or the server \citep{dai_federated_2020, zhou_flora_2021}. These methods work well when there are a relatively few number of clients and each client has adequate data to perform both the training and validation required for local tuning. 

Unlike prior work in federated HP tuning, we do not focus on modifications that interleave model training with HP optimization. Instead we point out that heterogeneous data distributed across a federated network results in noisy evaluations of the same model and attempt to isolate the impact of noisy evaluations on the HP search procedure. We show that under realistic constraints, the seemingly simple task of evaluating a global configuration poses challenges which have not received sufficient attention.

\vspace{-.1in}
\paragraph{Noisy hyperparameter tuning.}
Noisy evaluation can also be problematic in centralized hyperparameter tuning due to randomness in the training process. Most HP tuning algorithms do not explicitly consider noise, and simple tricks such as sampling more or resampling previously seen configurations \citep{hertel_quantity_2020} vary in effectiveness.

Bayesian optimization (BO) is a class of methods for sample-efficient optimization \citep{frazier_tutorial_2018}. Perhaps the most widely-used BO method is the expected improvement (EI) criterion \citep{snoek_practical_2012,mockus_application_1978}. TPE uses kernel density estimation to model HP quality and optimizes EI to select candidate points. However, in its naive form, EI assumes noiseless evaluations and is known to suffer in the presence of noise (see, e.g., S6.2 of \citet{balandat_botorch_2020}). Alternative BO approaches that pay attention to noisy evaluation include the \emph{knowledge gradient} \citep{frazier_knowledge-gradient_2008} and \emph{noisy expected improvement} \citep{letham_constrained_2019}. A main drawback shared by both methods is that they are more computationally expensive than EI and do not scale well to settings where high parallelism is desired.

Rather than viewing noise as an issue, multi-fidelity HPO methods improve efficiency by purposefully using cheap but noisy and/or biased evaluations to inform HP selection. Such methods limit the number of iterations \citep{li_hyperband_2017, falkner_bohb_2018}, dataset size \citep{klein_fast_2017}, or both \citep{wu_practical_2020} that are used to train a model. However, these approaches rely on the ability to also evaluate the highest fidelity setting (e.g., low noise or zero bias), which is not always possible in the context of FL. In addition, correctly modeling the impact of a low fidelity evaluation on the optimal configuration at the highest fidelity requires optimizing a one-step ``lookahead'' acquisition function and quickly becomes computationally expensive \citep{poloczek_multi-information_2017, wu_practical_2020}. As these methods (e.g,. Hyperband and BOHB) already rely on noise to improve efficiency, we suspect the additional noise from noisy evaluation saturates the methods and is a major reason for the poor performance of these approaches in our experiments.

\vspace{-.1in}
\paragraph{Private hyperparameter tuning.} Enforcing differential privacy~\cite{dwork_algorithmic_2013} requires adding randomization to the hyperparameter tuning process.
In this paper, our focus is not on developing new algorithms for private hyperparameter tuning but on the impact of randomness on the performance of tuning algorithms,
so we focus on a straightforward implementation of differential privacy.
However, there are a number of prior papers that explore more sophisticated algorithms for private hyperparameter tuning. 
One line of work ~\cite{chaudhuri_differentially_2011, liu_private_2019, papernot_hyperparameter_2022} focuses on efficient private selection from a discrete set of configurations.
\citet{chaudhuri2013stability} design an efficient procedure for hyperparameter selection under a \emph{stability} assumption on the scoring function.
Further works \citep{kusner_differentially_2015, dai_differentially_2021} develop
differentially private versions of Bayesian optimization to handle hyperparameter tuning.

\vspace{-.1in}
\section{Discussion \& Future Work}
\label{sec:discussion}
As we have shown, realistic FL settings present several sources of evaluation noise which can severely impact HP tuning methods. Our work highlights several best practices to mitigate the effects of noisy evaluation:

\vspace{-.1in}
\begin{enumerate}[leftmargin=*]
\itemsep0em 
    \item \textit{Use simple baselines.} Noisy evaluation can harm more sophisticated methods which perform early stopping or model the HP space.
    \item \textit{Obtain sufficiently large subsamples of validation clients.} For the datasets we consider, $\sim$100 is a reasonable number for non-private evaluation. However, these requirements grow with heterogeneity and privacy.
    \item \textit{Evaluate as representative a set of clients as possible.} Biased selection can lead to catastrophic drops in performance when client data is heterogeneous.
    \item \textit{Consider tuning HPs on proxy data.} If significant noise is expected, proxy data may be the most practical approach. As we show, even seemingly unrelated proxy data can be effective in high-noise regimes.
\end{enumerate}


\vspace{-.15in}
Beyond these key take-aways, our work also identifies several areas of future work, which we describe below.

\vspace{-.1in}
\paragraph{Early stopping in FL.} In the resource-constrained context of FL, early stopping methods are highly desirable for efficient HP tuning. However, as we show in this work, noisy evaluation can ruin the performance of these algorithms. We suspect this is due to the fact that these approaches (e.g. Hyperband, BOHB) already introduce their own source of noise to improve efficiency.
Therefore, a promising direction may be to extend early stopping methods to handle additional sources of noise or tailor them to federated settings. 

\vspace{-.1in}
\paragraph{Noisy BO.} Another future direction is considering `noisy BO' techniques
such as KG and NEI in the federated setting. One challenge to overcome is selecting a surrogate model that is able to accommodate the high levels of noise that we observe in FL. Another is that these acquisition functions are expensive to optimize: for KG, the time to suggest new configurations can be on the order of several minutes \citep{balandat_botorch_2020}. Depending on the relative time needed to evaluate a particular configuration, this can introduce a computational bottleneck on the server side.

\vspace{-.1in}
\paragraph{Resource-Aware HP Tuning.} More generally, the number of evaluations (e.g. early stopping) or the server-side overhead (e.g. BO) can significantly vary across methods, highlighting the need for resource-aware comparisons between complex tuning methods. Further, as resource constraints can vary across FL systems, a direction for future study is designing HP tuning methods which are aware of resource tradeoffs \citep{zhang_fedtune_2022}. In extreme cases, it would be beneficial to develop  FL methods that avoid or reduce the need for HP tuning at all~\cite{kairouz_advances_2021}.

\vspace{-.1in}
\paragraph{Heterogeneity-Aware HP Tuning.} While we model system heterogeneity with biased client sampling, more refined models should account for the inter-dependence of data and system heterogeneity \citep{maeng2022towards}. 
Although we have focused on the effects of HP tuning on average performance, it would be useful to explore the effect of heterogeneity in HP evaluation on tail performance as well, mirroring work in fair federated training~\cite{mohri2019agnostic,li2019fair}.

\vspace{-.1in}
\paragraph{Tuning via Proxy Data.} Finally, a key takeaway from our experiments is that proxy data (even seemingly unrelated) can be useful when faced with high-noise evaluation. However, it would be useful to further study this area to develop tools for easily determining if/when proxy data is appropriate.
Our work also suggests that public data, used to improve private
training of large models~\cite{li2021large, yu2021differentially, de2022unlocking},
may also be useful to improve private evaluation in a similar way.

\pagebreak

\section*{Acknowledgements}

We thank Mike Rabbat, Carole-Jean Wu, Hongyuan Zhan, Ilya Mironov, Liam Li, Ken Liu, Oscar Li, and Michael Kuchnik for their helpful comments. This work was supported in part by the National Science Foundation grants IIS1705121, IIS1838017, IIS2046613, IIS2112471, and funding from Meta, Morgan Stanley, Amazon, and Google. Any opinions, findings and conclusions or recommendations expressed in this material are those of the author(s) and do not necessarily reflect the views of any of these funding agencies.

\bibliography{references}

\begin{thebibliography}{57}
\providecommand{\natexlab}[1]{#1}
\providecommand{\url}[1]{\texttt{#1}}
\expandafter\ifx\csname urlstyle\endcsname\relax
  \providecommand{\doi}[1]{doi: #1}\else
  \providecommand{\doi}{doi: \begingroup \urlstyle{rm}\Url}\fi

\bibitem[Abadi et~al.(2016)Abadi, Chu, Goodfellow, McMahan, Mironov, Talwar,
  and Zhang]{abadi_deep_2016}
Abadi, M., Chu, A., Goodfellow, I., McMahan, H.~B., Mironov, I., Talwar, K.,
  and Zhang, L.
\newblock Deep {Learning} with {Differential} {Privacy}.
\newblock In \emph{{ACM} {SIGSAC} {Conference} on {Computer} and
  {Communications} {Security}}, pp.\  308--318, Vienna Austria, 2016.

\bibitem[Agrawal et~al.(2021)Agrawal, Sarkar, Alazab, Maddikunta, Gadekallu,
  and Pham]{agrawal_genetic_2021}
Agrawal, S., Sarkar, S., Alazab, M., Maddikunta, P. K.~R., Gadekallu, T.~R.,
  and Pham, Q.-V.
\newblock Genetic {CFL}: {Optimization} of {Hyper}-{Parameters} in {Clustered}
  {Federated} {Learning}.
\newblock \emph{Computational Intelligence and Neuroscience}, 2021, 2021.

\bibitem[Balandat et~al.(2020)Balandat, Karrer, Jiang, Daulton, Letham, Wilson,
  and Bakshy]{balandat_botorch_2020}
Balandat, M., Karrer, B., Jiang, D., Daulton, S., Letham, B., Wilson, A.~G.,
  and Bakshy, E.
\newblock Botorch: a framework for efficient monte-carlo bayesian optimization.
\newblock \emph{Advances in neural information processing systems},
  33:\penalty0 21524--21538, 2020.

\bibitem[Bergstra \& Bengio(2012)Bergstra and Bengio]{bergstra_random_2012}
Bergstra, J. and Bengio, Y.
\newblock Random {Search} for {Hyper}-{Parameter} {Optimization}.
\newblock \emph{Journal of Machine Learning Research}, 13\penalty0
  (10):\penalty0 281--305, 2012.

\bibitem[Bergstra et~al.(2011)Bergstra, Bardenet, Bengio, and
  Kégl]{bergstra_algorithms_2011}
Bergstra, J., Bardenet, R., Bengio, Y., and Kégl, B.
\newblock Algorithms for {Hyper}-{Parameter} {Optimization}.
\newblock In \emph{Advances in {Neural} {Information} {Processing} {Systems}}.
  Curran Associates, Inc., 2011.

\bibitem[Bonawitz et~al.(2019)Bonawitz, Eichner, Grieskamp, Huba, Ingerman,
  Ivanov, Kiddon, Kone{\v{c}}n{\'y}, Mazzocchi, McMahan, Van~Overveldt, Petrou,
  Ramage, and Roselander]{bonawitz_towards_2019}
Bonawitz, K., Eichner, H., Grieskamp, W., Huba, D., Ingerman, A., Ivanov, V.,
  Kiddon, C., Kone{\v{c}}n{\'y}, J., Mazzocchi, S., McMahan, B., Van~Overveldt,
  T., Petrou, D., Ramage, D., and Roselander, J.
\newblock Towards {Federated} {Learning} at {Scale}: {System} {Design}.
\newblock \emph{Proceedings of Machine Learning and Systems}, 2019.

\bibitem[Caldas et~al.(2018)Caldas, Duddu, Wu, Li, Kone{\v{c}}n{\'y}, McMahan,
  Smith, and Talwalkar]{caldas_leaf_2018}
Caldas, S., Duddu, S. M.~K., Wu, P., Li, T., Kone{\v{c}}n{\'y}, J., McMahan,
  H.~B., Smith, V., and Talwalkar, A.
\newblock Leaf: A benchmark for federated settings.
\newblock \emph{arXiv preprint arXiv:1812.01097}, 2018.

\bibitem[Charles et~al.(2021)Charles, Garrett, Huo, Shmulyian, and
  Smith]{charles_large-cohort_2021}
Charles, Z., Garrett, Z., Huo, Z., Shmulyian, S., and Smith, V.
\newblock On {Large}-{Cohort} {Training} for {Federated} {Learning}.
\newblock In \emph{Advances in {Neural} {Information} {Processing} {Systems}},
  2021.

\bibitem[Chaudhuri \& Vinterbo(2013)Chaudhuri and
  Vinterbo]{chaudhuri2013stability}
Chaudhuri, K. and Vinterbo, S.~A.
\newblock A stability-based validation procedure for differentially private
  machine learning.
\newblock \emph{Advances in Neural Information Processing Systems}, 26, 2013.

\bibitem[Chaudhuri et~al.(2011)Chaudhuri, Monteleoni, and
  Sarwate]{chaudhuri_differentially_2011}
Chaudhuri, K., Monteleoni, C., and Sarwate, A.~D.
\newblock Differentially private empirical risk minimization.
\newblock \emph{Journal of Machine Learning Research}, 12\penalty0 (3), 2011.

\bibitem[Chen et~al.(2023)Chen, Wang, Chen, Yan, and Lin]{chen_dap_2023}
Chen, Q., Wang, Z., Chen, J., Yan, H., and Lin, X.
\newblock Dap-fl: Federated learning flourishes by adaptive tuning and secure
  aggregation.
\newblock \emph{IEEE Transactions on Parallel and Distributed Systems}, 2023.

\bibitem[Dai et~al.(2020)Dai, Low, and Jaillet]{dai_federated_2020}
Dai, Z., Low, B. K.~H., and Jaillet, P.
\newblock Federated {Bayesian} {Optimization} via {Thompson} {Sampling}.
\newblock In \emph{Advances in {Neural} {Information} {Processing} {Systems}},
  2020.

\bibitem[Dai et~al.(2021)Dai, Low, and Jaillet]{dai_differentially_2021}
Dai, Z., Low, B. K.~H., and Jaillet, P.
\newblock Differentially private federated {Bayesian} optimization with
  distributed exploration.
\newblock \emph{Advances in Neural Information Processing Systems}, 2021.

\bibitem[De et~al.(2022)De, Berrada, Hayes, Smith, and Balle]{de2022unlocking}
De, S., Berrada, L., Hayes, J., Smith, S.~L., and Balle, B.
\newblock Unlocking high-accuracy differentially private image classification
  through scale.
\newblock \emph{arXiv preprint arXiv:2204.13650}, 2022.

\bibitem[Dwork \& Roth(2013)Dwork and Roth]{dwork_algorithmic_2013}
Dwork, C. and Roth, A.
\newblock The {Algorithmic} {Foundations} of {Differential} {Privacy}.
\newblock \emph{Foundations and Trends in Theoretical Computer Science},
  9\penalty0 (3-4):\penalty0 211--407, 2013.

\bibitem[Falkner et~al.(2018)Falkner, Klein, and Hutter]{falkner_bohb_2018}
Falkner, S., Klein, A., and Hutter, F.
\newblock {BOHB}: {Robust} and {Efficient} {Hyperparameter} {Optimization} at
  {Scale}.
\newblock In \emph{{International} {Conference} on {Machine} {Learning}}, 2018.

\bibitem[Frazier(2018)]{frazier_tutorial_2018}
Frazier, P.~I.
\newblock A tutorial on bayesian optimization.
\newblock \emph{arXiv preprint arXiv:1807.02811}, 2018.

\bibitem[Frazier et~al.(2008)Frazier, Powell, and
  Dayanik]{frazier_knowledge-gradient_2008}
Frazier, P.~I., Powell, W.~B., and Dayanik, S.
\newblock A {Knowledge}-{Gradient} {Policy} for {Sequential} {Information}
  {Collection}.
\newblock \emph{SIAM Journal on Control and Optimization}, 47\penalty0
  (5):\penalty0 2410--2439, 2008.

\bibitem[Hertel et~al.(2020)Hertel, Baldi, and Gillen]{hertel_quantity_2020}
Hertel, L., Baldi, P., and Gillen, D.~L.
\newblock Quantity vs. quality: On hyperparameter optimization for deep
  reinforcement learning.
\newblock \emph{arXiv preprint arXiv:2007.14604}, 2020.

\bibitem[Holly et~al.(2022)Holly, Hiessl, Lakani, Schall, Heitzinger, and
  Kemnitz]{holly_evaluation_2021}
Holly, S., Hiessl, T., Lakani, S.~R., Schall, D., Heitzinger, C., and Kemnitz,
  J.
\newblock Evaluation of hyperparameter-optimization approaches in an industrial
  federated learning system.
\newblock In \emph{Data Science--Analytics and Applications: Proceedings of the
  4th International Data Science Conference--iDSC2021}, pp.\  6--13. Springer,
  2022.

\bibitem[Hsu et~al.(2019)Hsu, Qi, and Brown]{hsu_measuring_2019}
Hsu, T.-M.~H., Qi, H., and Brown, M.
\newblock Measuring the effects of non-identical data distribution for
  federated visual classification.
\newblock \emph{arXiv preprint arXiv:1909.06335}, 2019.

\bibitem[Jaderberg et~al.(2017)Jaderberg, Dalibard, Osindero, Czarnecki,
  Donahue, Razavi, Vinyals, Green, Dunning, Simonyan,
  et~al.]{jaderberg_population_2017}
Jaderberg, M., Dalibard, V., Osindero, S., Czarnecki, W.~M., Donahue, J.,
  Razavi, A., Vinyals, O., Green, T., Dunning, I., Simonyan, K., et~al.
\newblock Population based training of neural networks.
\newblock \emph{arXiv preprint arXiv:1711.09846}, 2017.

\bibitem[Kairouz et~al.(2021)Kairouz, McMahan, Avent, Bellet, Bennis, Bhagoji,
  Bonawitz, Charles, Cormode, Cummings, et~al.]{kairouz_advances_2021}
Kairouz, P., McMahan, H.~B., Avent, B., Bellet, A., Bennis, M., Bhagoji, A.~N.,
  Bonawitz, K., Charles, Z., Cormode, G., Cummings, R., et~al.
\newblock Advances and open problems in federated learning.
\newblock \emph{Foundations and Trends{\textregistered} in Machine Learning},
  14\penalty0 (1--2):\penalty0 1--210, 2021.

\bibitem[Khodak et~al.(2021)Khodak, Tu, Li, Li, Balcan, Smith, and
  Talwalkar]{khodak_federated_2021}
Khodak, M., Tu, R., Li, T., Li, L., Balcan, M.-F.~F., Smith, V., and Talwalkar,
  A.
\newblock Federated {Hyperparameter} {Tuning}: {Challenges}, {Baselines}, and
  {Connections} to {Weight}-{Sharing}.
\newblock In \emph{Advances in {Neural} {Information} {Processing} {Systems}},
  2021.

\bibitem[Klein et~al.(2017)Klein, Falkner, Bartels, Hennig, and
  Hutter]{klein_fast_2017}
Klein, A., Falkner, S., Bartels, S., Hennig, P., and Hutter, F.
\newblock Fast {Bayesian} {Optimization} of {Machine} {Learning}
  {Hyperparameters} on {Large} {Datasets}.
\newblock In \emph{{International} {Conference} on {Artificial} {Intelligence}
  and {Statistics}}. PMLR, 2017.

\bibitem[Koskela \& Honkela(2018)Koskela and Honkela]{koskela_learning_2018}
Koskela, A. and Honkela, A.
\newblock Learning rate adaptation for federated and differentially private
  learning.
\newblock \emph{arXiv preprint arXiv:1809.03832}, 2018.

\bibitem[Krizhevsky \& Hinton(2009)Krizhevsky and
  Hinton]{krizhevsky_learning_2009}
Krizhevsky, A. and Hinton, G.
\newblock Learning multiple layers of features from tiny images.
\newblock Technical Report~0, University of Toronto, Toronto, Ontario, 2009.

\bibitem[Kuo(2023)]{kevin_kuo_2023_7819606}
Kuo, K.
\newblock {imkevinkuo/noisy-eval-in-fl: Release: MLSys'23 Artifact Evaluation},
  April 2023.
\newblock URL \url{https://doi.org/10.5281/zenodo.7819606}.

\bibitem[Kusner et~al.(2015)Kusner, Gardner, Garnett, and
  Weinberger]{kusner_differentially_2015}
Kusner, M., Gardner, J., Garnett, R., and Weinberger, K.
\newblock Differentially {Private} {Bayesian} {Optimization}.
\newblock In \emph{{International} {Conference} on {Machine} {Learning}}, 2015.

\bibitem[Letham et~al.(2019)Letham, Karrer, Ottoni, and
  Bakshy]{letham_constrained_2019}
Letham, B., Karrer, B., Ottoni, G., and Bakshy, E.
\newblock Constrained {Bayesian} {Optimization} with {Noisy} {Experiments}.
\newblock \emph{Bayesian Analysis}, 14\penalty0 (2):\penalty0 495--519, 2019.

\bibitem[Li et~al.(2017)Li, Jamieson, DeSalvo, Rostamizadeh, and
  Talwalkar]{li_hyperband_2017}
Li, L., Jamieson, K., DeSalvo, G., Rostamizadeh, A., and Talwalkar, A.
\newblock Hyperband: {A} novel bandit-based approach to hyperparameter
  optimization.
\newblock \emph{Journal of Machine Learning Research}, 18\penalty0
  (1):\penalty0 6765--6816, 2017.

\bibitem[Li et~al.(2020{\natexlab{a}})Li, Sahu, Talwalkar, and
  Smith]{li_federated_2020}
Li, T., Sahu, A.~K., Talwalkar, A., and Smith, V.
\newblock Federated {Learning}: {Challenges}, {Methods}, and {Future}
  {Directions}.
\newblock \emph{IEEE Signal Processing Magazine}, 37:\penalty0 50--60,
  2020{\natexlab{a}}.

\bibitem[Li et~al.(2020{\natexlab{b}})Li, Sahu, Zaheer, Sanjabi, Talwalkar, and
  Smith]{li_federated_2020-1}
Li, T., Sahu, A.~K., Zaheer, M., Sanjabi, M., Talwalkar, A., and Smith, V.
\newblock Federated optimization in heterogeneous networks.
\newblock \emph{Proceedings of Machine Learning and Systems},
  2020{\natexlab{b}}.

\bibitem[Li et~al.(2020{\natexlab{c}})Li, Sanjabi, Beirami, and
  Smith]{li2019fair}
Li, T., Sanjabi, M., Beirami, A., and Smith, V.
\newblock Fair resource allocation in federated learning.
\newblock In \emph{International Conference on Learning Representations},
  2020{\natexlab{c}}.

\bibitem[Li et~al.(2022)Li, Tramer, Liang, and Hashimoto]{li2021large}
Li, X., Tramer, F., Liang, P., and Hashimoto, T.
\newblock Large language models can be strong differentially private learners.
\newblock In \emph{International Conference on Learning Representations}, 2022.

\bibitem[Liu \& Talwar(2019)Liu and Talwar]{liu_private_2019}
Liu, J. and Talwar, K.
\newblock Private selection from private candidates.
\newblock In \emph{{ACM} {SIGACT} {Symposium} on {Theory} of {Computing}},
  2019.

\bibitem[Maeng et~al.(2022)Maeng, Lu, Melis, Nguyen, Rabbat, and
  Wu]{maeng2022towards}
Maeng, K., Lu, H., Melis, L., Nguyen, J., Rabbat, M., and Wu, C.-J.
\newblock Towards fair federated recommendation learning: Characterizing the
  inter-dependence of system and data heterogeneity.
\newblock In \emph{Proceedings of the 16th ACM Conference on Recommender
  Systems}, RecSys '22, pp.\  156–167. Association for Computing Machinery,
  2022.

\bibitem[McMahan et~al.(2017)McMahan, Moore, Ramage, Hampson, and
  Arcas]{mcmahan_communication-efficient_2017}
McMahan, B., Moore, E., Ramage, D., Hampson, S., and Arcas, B. A.~y.
\newblock Communication-{Efficient} {Learning} of {Deep} {Networks} from
  {Decentralized} {Data}.
\newblock In \emph{{International} {Conference} on {Artificial} {Intelligence}
  and {Statistics}}. PMLR, 2017.

\bibitem[McMahan et~al.(2018)McMahan, Ramage, Talwar, and
  Zhang]{brendan2018learning}
McMahan, H.~B., Ramage, D., Talwar, K., and Zhang, L.
\newblock Learning differentially private recurrent language models.
\newblock In \emph{International Conference on Learning Representations}, 2018.

\bibitem[Mlodozeniec et~al.(2023)Mlodozeniec, Reisser, and
  Louizos]{mlodozeniechyperparameter}
Mlodozeniec, B.~K., Reisser, M., and Louizos, C.
\newblock Hyperparameter optimization through neural network partitioning.
\newblock In \emph{The Eleventh International Conference on Learning
  Representations}, 2023.

\bibitem[Mockus et~al.(1978)Mockus, Tiesis, and
  Zilinskas]{mockus_application_1978}
Mockus, J., Tiesis, V., and Zilinskas, A.
\newblock The application of {Bayesian} methods for seeking the extremum.
\newblock \emph{Towards global optimization}, 2\penalty0 (117-129):\penalty0 2,
  1978.

\bibitem[Mohri et~al.(2019)Mohri, Sivek, and Suresh]{mohri2019agnostic}
Mohri, M., Sivek, G., and Suresh, A.~T.
\newblock Agnostic federated learning.
\newblock In \emph{International Conference on Machine Learning}, 2019.

\bibitem[Mostafa(2019)]{mostafa_robust_2019}
Mostafa, H.
\newblock Robust federated learning through representation matching and
  adaptive hyper-parameters.
\newblock \emph{arXiv preprint arXiv:1912.13075}, 2019.

\bibitem[Papernot \& Steinke(2022)Papernot and
  Steinke]{papernot_hyperparameter_2022}
Papernot, N. and Steinke, T.
\newblock Hyperparameter tuning with {Renyi} differential privacy.
\newblock In \emph{International Conference on Learning Representations}, 2022.

\bibitem[Poloczek et~al.(2017)Poloczek, Wang, and
  Frazier]{poloczek_multi-information_2017}
Poloczek, M., Wang, J., and Frazier, P.
\newblock Multi-{Information} {Source} {Optimization}.
\newblock In \emph{Advances in {Neural} {Information} {Processing} {Systems}},
  2017.

\bibitem[Qiao et~al.(2021)Qiao, Su, and Zhang]{qiao_oneshot_2021}
Qiao, G., Su, W., and Zhang, L.
\newblock Oneshot differentially private top-k selection.
\newblock In \emph{International Conference on Machine Learning}, pp.\
  8672--8681. PMLR, 2021.

\bibitem[Radford et~al.(2019)Radford, Wu, Child, Luan, Amodei, and
  Sutskever]{radford_language_2019}
Radford, A., Wu, J., Child, R., Luan, D., Amodei, D., and Sutskever, I.
\newblock Language models are unsupervised multitask learners.
\newblock 2019.

\bibitem[Reddi et~al.(2020)Reddi, Charles, Zaheer, Garrett, Rush,
  Kone{\v{c}}n{\'y}, Kumar, and McMahan]{reddi_adaptive_2020}
Reddi, S.~J., Charles, Z., Zaheer, M., Garrett, Z., Rush, K.,
  Kone{\v{c}}n{\'y}, J., Kumar, S., and McMahan, H.~B.
\newblock Adaptive {Federated} {Optimization}.
\newblock In \emph{International {Conference} on {Learning} {Representations}},
  2020.

\bibitem[Snoek et~al.(2012)Snoek, Larochelle, and Adams]{snoek_practical_2012}
Snoek, J., Larochelle, H., and Adams, R.~P.
\newblock Practical {Bayesian} optimization of machine learning algorithms.
\newblock In \emph{Advances in {Neural} {Information} {Processing} {Systems}},
  2012.

\bibitem[{The TensorFlow Federated Authors}(2019)]{authors_tensorflow_2019}
{The TensorFlow Federated Authors}.
\newblock {TensorFlow} {Federated} {Stack} {Overflow} dataset, 2019.
\newblock URL
  \url{https://github.com/google/fedjax/blob/main/fedjax/datasets/stackoverflow.py#L71}.

\bibitem[Wang et~al.(2021)Wang, Charles, Xu, Joshi, McMahan, Al-Shedivat,
  Andrew, Avestimehr, Daly, Data, et~al.]{wang_field_2021}
Wang, J., Charles, Z., Xu, Z., Joshi, G., McMahan, H.~B., Al-Shedivat, M.,
  Andrew, G., Avestimehr, S., Daly, K., Data, D., et~al.
\newblock A field guide to federated optimization.
\newblock \emph{arXiv preprint arXiv:2107.06917}, 2021.

\bibitem[Wang et~al.(2022)Wang, Kuang, Zhang, Ding, and Li]{wang_fedhpo-b_2022}
Wang, Z., Kuang, W., Zhang, C., Ding, B., and Li, Y.
\newblock Fedhpo-b: A benchmark suite for federated hyperparameter
  optimization.
\newblock \emph{arXiv preprint arXiv:2206.03966}, 2022.

\bibitem[Wu et~al.(2020)Wu, Toscano-Palmerin, Frazier, and
  Wilson]{wu_practical_2020}
Wu, J., Toscano-Palmerin, S., Frazier, P.~I., and Wilson, A.~G.
\newblock Practical {Multi}-fidelity {Bayesian} {Optimization} for
  {Hyperparameter} {Tuning}.
\newblock In \emph{{Uncertainty} in {Artificial} {Intelligence} {Conference}},
  2020.

\bibitem[Yu et~al.(2021)Yu, Naik, Backurs, Gopi, Inan, Kamath, Kulkarni, Lee,
  Manoel, Wutschitz, et~al.]{yu2021differentially}
Yu, D., Naik, S., Backurs, A., Gopi, S., Inan, H.~A., Kamath, G., Kulkarni, J.,
  Lee, Y.~T., Manoel, A., Wutschitz, L., et~al.
\newblock Differentially private fine-tuning of language models.
\newblock In \emph{International Conference on Learning Representations}, 2021.

\bibitem[Yuan et~al.(2022)Yuan, Morningstar, Ning, and Singhal]{yuan_what_2022}
Yuan, H., Morningstar, W.~R., Ning, L., and Singhal, K.
\newblock What do we mean by generalization in federated learning?
\newblock In \emph{International Conference on Learning Representations}, 2022.

\bibitem[Zhang et~al.(2022)Zhang, Zhang, Liu, Mohapatra, and
  DeLucia]{zhang_fedtune_2022}
Zhang, H., Zhang, M., Liu, X., Mohapatra, P., and DeLucia, M.
\newblock Fedtune: Automatic tuning of federated learning hyper-parameters from
  system perspective.
\newblock In \emph{MILCOM 2022-2022 IEEE Military Communications Conference
  (MILCOM)}, pp.\  478--483. IEEE, 2022.

\bibitem[Zhou et~al.(2021)Zhou, Ram, Salonidis, Baracaldo, Samulowitz, and
  Ludwig]{zhou_flora_2021}
Zhou, Y., Ram, P., Salonidis, T., Baracaldo, N., Samulowitz, H., and Ludwig, H.
\newblock Flora: Single-shot hyper-parameter optimization for federated
  learning.
\newblock \emph{arXiv preprint arXiv:2112.08524}, 2021.

\end{thebibliography}
\bibliographystyle{mlsys2023}

\clearpage
\appendix
\section{Method Details.}
\label{sec:appendix:hpo}
\paragraph{HP Tuning Methods.} We consider two classes of methods: \emph{model-free} and \emph{model-based}. Random and grid search are examples of the simplest model-free methods which do not make any assumptions about the function being optimized besides the HP space to search over. To generate candidate configurations, grid search discretizes the hyperparameter space into a multi-dimensional grid, while random search samples hyperparameter values from a predefined distribution, typically discrete or (log-)uniform/normal. Both methods sample a set of candidates in an iid fashion, evaluate them, and return the best-performing configuration. More complex examples of model-free methods include Hyperband (HB) \citep{li_hyperband_2017} and Population-Based Training \citep{jaderberg_population_2017}.

Model-based methods iterate between fitting a \textit{surrogate} model of the hyperparameter response function $f(\theta)$ on previously tested configurations and selecting the next query configuration $\theta^*$ by optimizing some criterion on the current surrogate. A classic instantiation selects $\theta^*$ by optimizing \textit{expected improvement} on a \textit{Gaussian process} (GP) surrogate model. The \emph{tree-structured Parzen estimator} (TPE) is an alternative model that has been shown to outperform GPs in certain cases \citep{bergstra_algorithms_2011}. Finally, it is also possible to combine model-based methods with the early stopping techniques in model-free methods. For example, BOHB uses TPE to select candidate configurations for Hyperband \citep{falkner_bohb_2018}. 
We now describe HB, TPE, and BOHB as we use them in our experiments. 

\textbf{HB} is an extension of random search which eliminates poorly-performing configurations early in training, allowing more resources to be allocated on promising configurations. A subroutine called Successive Halving (SHA) performs the eliminations; it takes as input $n$ configurations, an elimination rate $\eta$ (typically set to $\eta=3$), and a minimum resource $r_0$. After training all $n$ configurations for $r_0$ iterations, SHA eliminates all but the top $\lfloor n/\eta \rfloor$ configurations and scales up their resource budgets $r_{i+1} = r_i\eta$. This step repeats until less than $\eta$ configurations remain. Hyperband can be described as a wrapper algorithm  which runs multiple configurations of $\text{SHA}(n,\eta,r_0)$ to balance between exploration (partially training many configurations) and exploitation (fully training a few configurations).

\textbf{TPE} models $p(\theta|y)$ with two densities $\ell(\theta)$ and $g(\theta)$:
\begin{align*}
p(\theta|y) = \left\{
\begin{array}{lr}
    \ell(\theta) & \text{if } y < y^*,\\
    g(\theta) & \text{if } y \geq  y^*
\end{array}\right.
\end{align*}
TPE splits the current observations $\{(\theta^{(i)}, y^{(i)})\}$ into two groups based on the threshold $y^*$: observations with $y^{(i)} < y^*$ are used to estimate $\ell(\theta)$ while those with $y^{(i)} \geq y^*$ are used to estimate $g(\theta)$. Optimizing EI for this model is equivalent to minimizing the quantity $g(x)/\ell(x)$, which is done by taking a minimum over random samples from $\ell(x)$.

\textbf{BOHB} replaces the default random sampling in HB with the TPE acquisition function. BOHB starts with random sampling, uses low-fidelity evaluations to form the TPE densities, and gradually switches to higher fidelity evaluations as they become available. 

\section{HP Search Space.}
\label{sec:appendix:hpspace}
Server (FedAdam) hyperparameters:
\begin{align*}
    \log_{10} \text{ lr } & : \text{Unif}[-6, -1] \\
    \text{(1\textsuperscript{st} moment decay) } \beta_1 & : \text{Unif}[0, 0.9] \\
    \text{(2\textsuperscript{nd} moment decay) } \beta_2 & : \text{Unif}[0, 0.999] \\
    \text{(lr\_decay) } \gamma & : 0.9999 \\
\end{align*}

Client (SGD) hyperparameters:
\begin{align*}
    \log_{10} \text{ lr } & : \text{Unif}[-6, 0] \\
    \text{momentum} & : \text{Unif}[0, 0.9] \\
    \text{weight\_decay} & : 0.00005 \\
    \text{batch\_size} & : [32, 64, 128] \\
    \text{epochs} & : 1
\end{align*}

\begin{figure}[h!]
\centering
    \includegraphics[width=8cm]{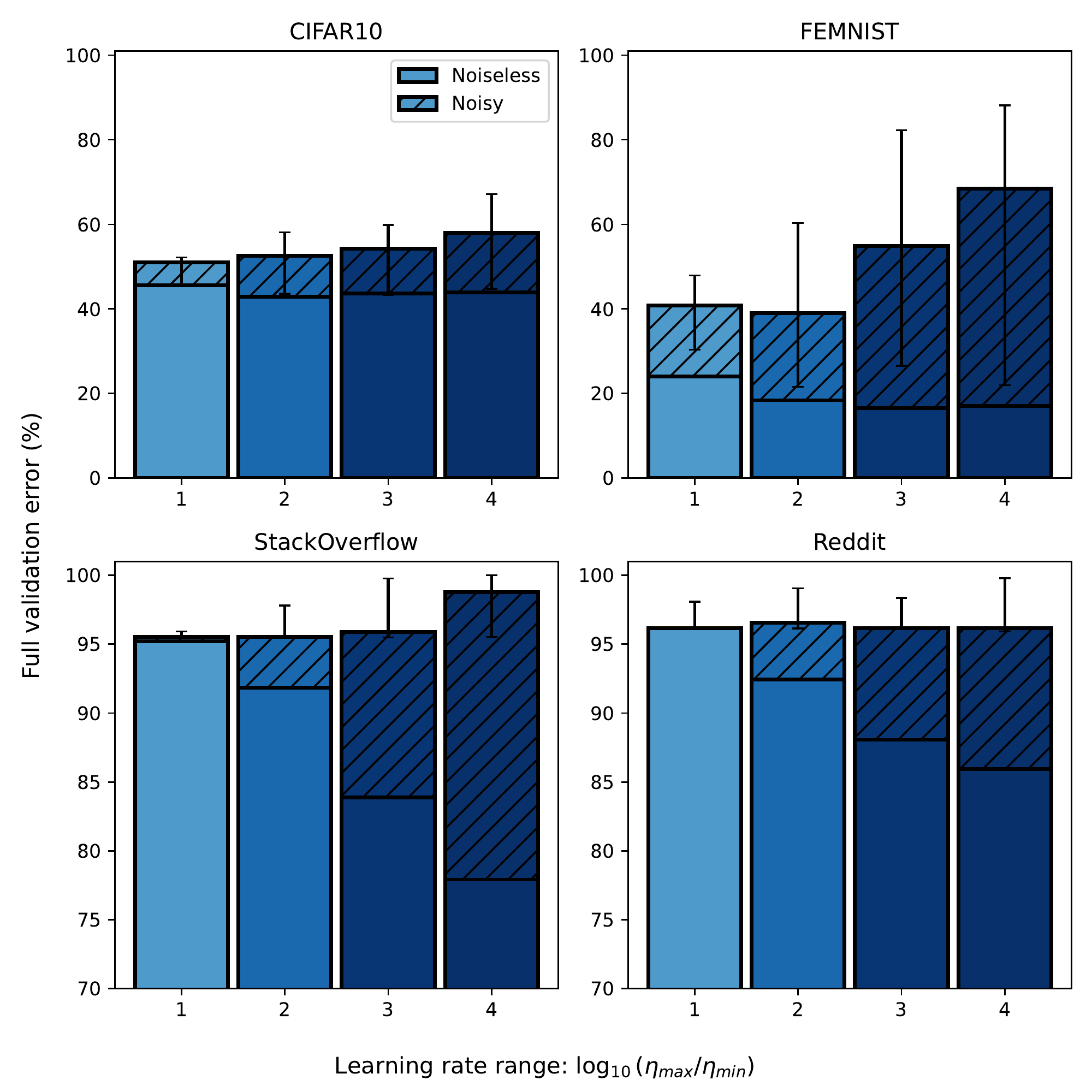}
    \caption{We run RS with a large budget ($K=128$) in a high-noise (1 client subsample, $\eps=10$ privacy) and noiseless setting. The x-axis varies (from left to right, smallest to largest) the \textit{range} of searched FedAdam server learning rates. Searching a larger HP space is beneficial in noiseless settings, but can counterintuitively harm performance in noisy settings.}
    \label{fig:appendix:hpo_space_exp}
\end{figure}

\clearpage

\section{Additional Tables and Figures}
\label{sec:appendix:figs}

\textbf{HP Space Experiment.} We include an additional experiment in Fig.~\ref{fig:appendix:hpo_space_exp} to investigate how the choice of HP space interacts with noisy evaluation. Intuitively, if there is a sufficient tuning budget, enlarging the search space offers more opportunities to improve performance (as long as the globally optimal HPs have not already been identified). However, depending on the selection of these HP spaces, this observation can be reversed when evaluation is noisy. We consider nested search intervals for the \textit{server learning rate}, which we observed to be the most sensitive HP. All other HP ranges match Appendix~\ref{sec:appendix:hpspace}. The search is centered on $10^{-3}$ and the range $[\eta_{min}, \eta_{max}]$ is adjusted such that $\log_{10}(\eta_{max}/\eta_{min}) \in \{1,2,3,4\}$. 1 is the smallest range ($[10^{-4.5},10^{-3.5}]$) while 4 is the largest ($[10^{-6},10^{-2}]$).

\begin{figure*}
\centering
    \includegraphics[width=6cm]{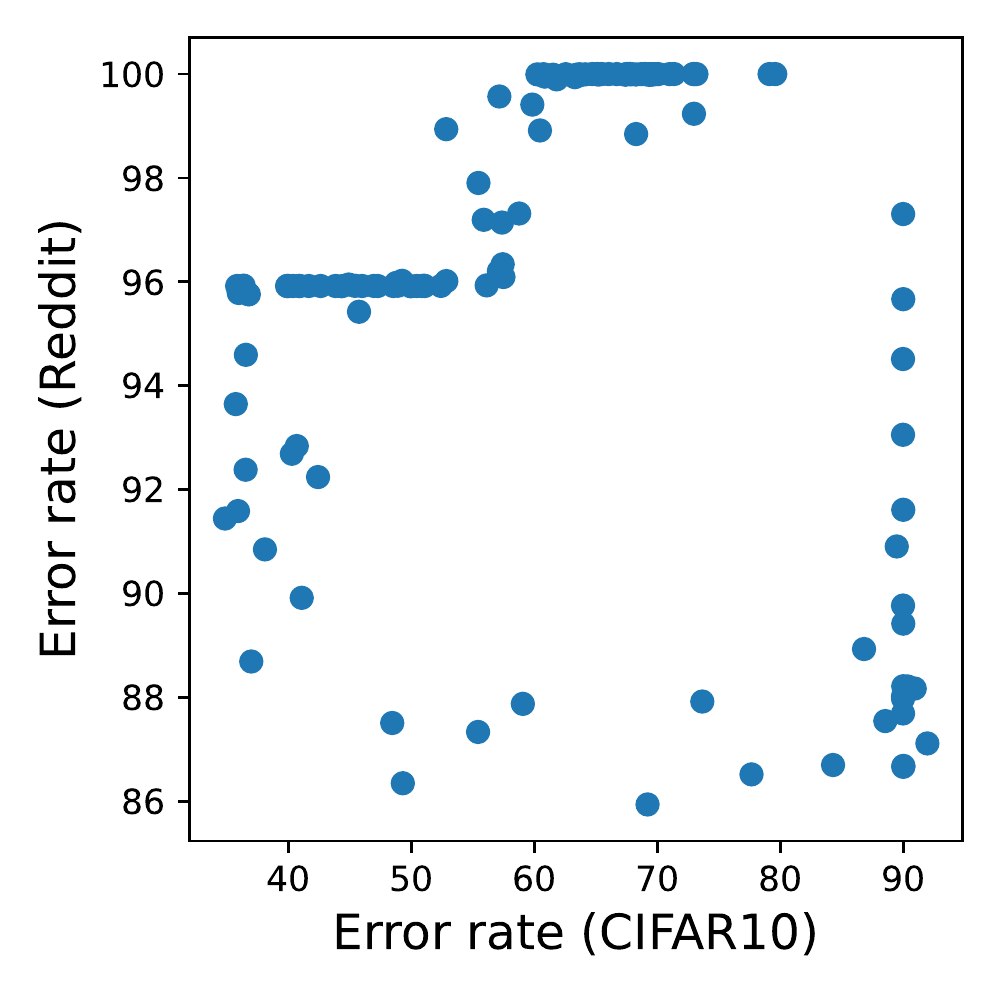}
    \hspace{1cm}
    \includegraphics[width=6cm]{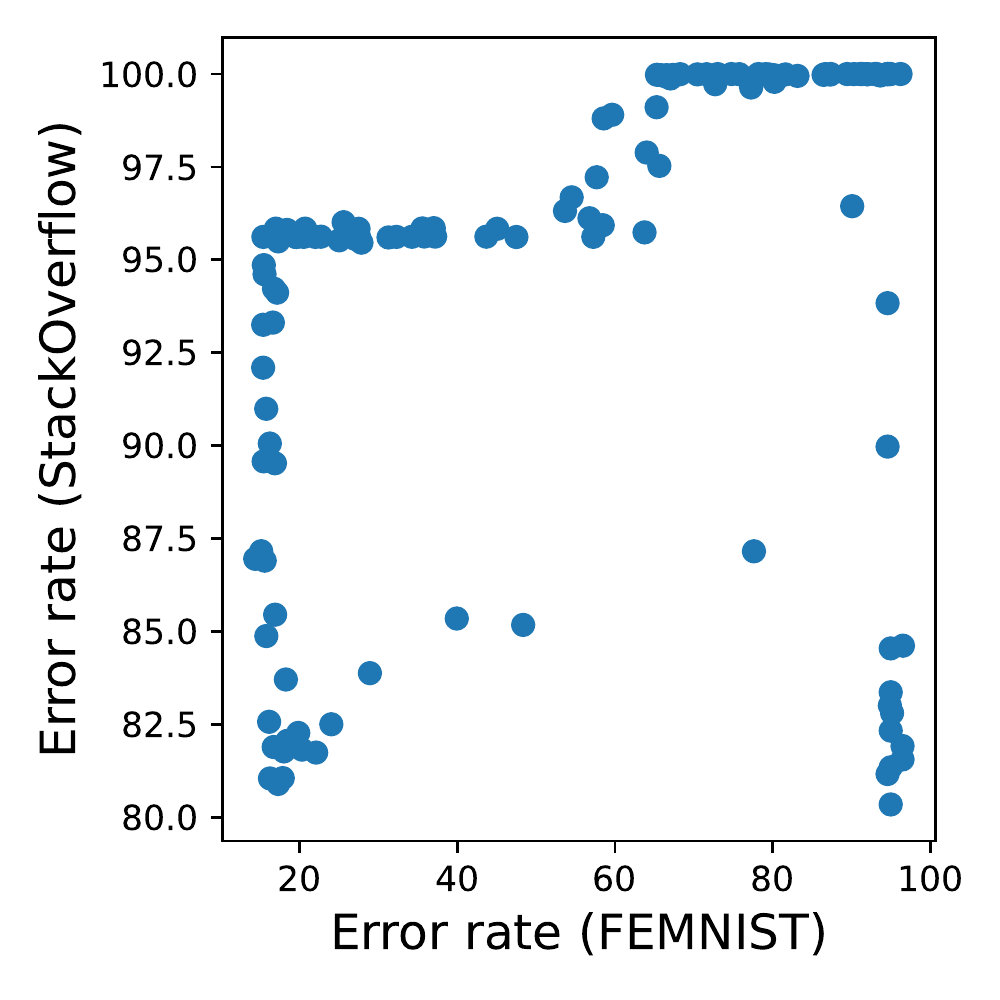}\caption{Error rate of configurations used to train separate models on CIFAR10/Reddit and FEMNIST/StackOverflow.}
    \label{fig:appendix:px_scatter}
\end{figure*}

\begin{figure*}
\centering
    \includegraphics[width=16cm]{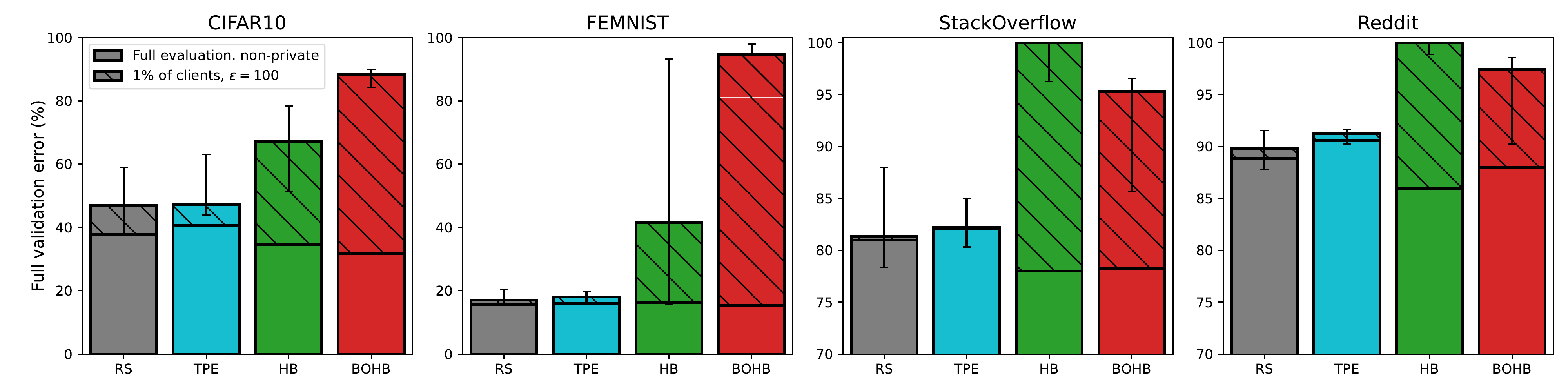}
    \caption{Performance of RS, HB, TPE, and BOHB at \textit{2000 allocated training rounds}. The hatched bars show degradation from two sources of noise (subsampling 1\% of clients and $\eps=100$ privacy).}
    \label{fig:appendix:hpo_bar_2k}
\end{figure*}

\begin{figure*}
\centering
    \includegraphics[width=16cm]{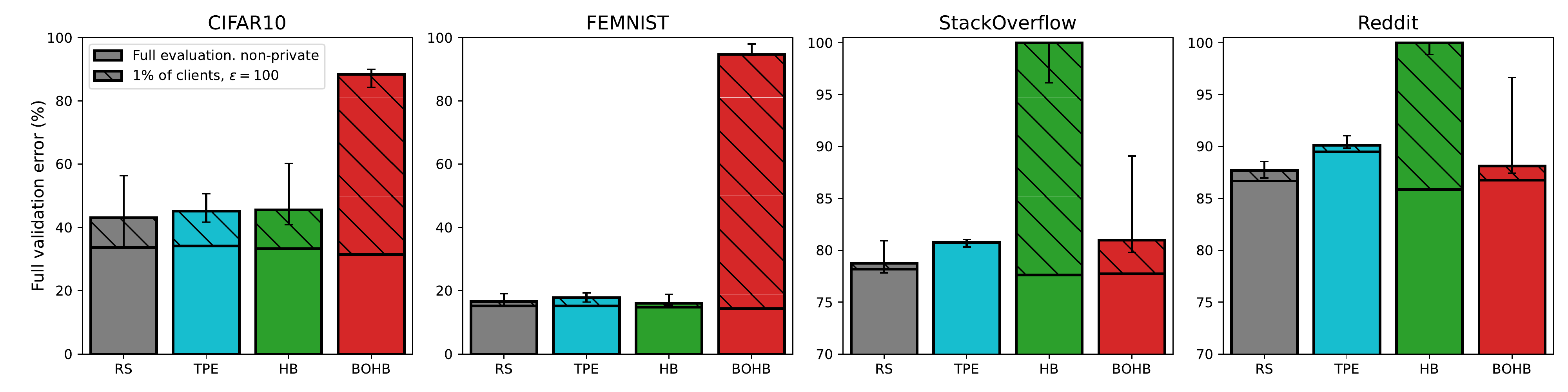}
    \caption{Identical to Figure~\ref{fig:appendix:hpo_bar_2k}, but at \textit{6480 allocated training rounds} (the full budget).}
    \label{fig:appendix:hpo_bar_max}
\end{figure*}

\textbf{Extra Figures.} Table~\ref{tab:appendix:datasets} shows additional dataset information. Figure~\ref{fig:appendix:px_scatter} shows configuration performance on the two dataset pairs not shown in Section~\ref{sec:proxy}. Figures~\ref{fig:appendix:hpo_bar_2k} and~\ref{fig:appendix:hpo_bar_max} compare the performance across HPO methods under subsampling and privacy noise. Figure~\ref{fig:appendix:hpo_bar_2k} shows performance when 1/3rd of the budget is used up, while Figure~\ref{fig:appendix:hpo_bar_max} shows performance at the full budget.

\begin{algorithm}[h!]
\caption{Random Search (RS) for centralized data}\label{alg:rs}
\begin{algorithmic}
\REQUIRE 
$\Theta$ (hyperparameter space)
\\ $K$ (num. of HP configs to search)
\\ $R$ (training epochs per HP config).
\\ $D_\text{tr}$ (training dataset)
\\ $D_\text{val}$ (validation dataset)
\\ $D_\text{test}$ (testing dataset)
\\ \texttt{OPT}() (optimization method e.g. SGD.)
\FOR {$k = 1, ..., K$}
    \STATE Sample HP config $\theta_k\sim\Theta$ uniformly at random.
    \STATE Initialize model parameters $w_k$.
    \STATE \color{teal}\textbf{\# Training}
    \color{black}
    \FOR {$r = 1, ..., R$}
        \color{teal}
        \STATE \#~\texttt{OPT}() trains $w_k$ for one `epoch'. 
        \STATE \#~Batching is handled by \texttt{OPT}() and $\theta_k$.
        \color{red}
        \STATE $w_k \gets$ \texttt{OPT}($w_k,D_\text{tr};\theta_k$) 
        \color{black}
    \ENDFOR
    \\ \color{teal} \textbf{\# Evaluation}
    \color{red}
    \STATE $L_k =$ Error rate of $w$ on $D_\text{val}$
    \color{black}
\ENDFOR
\STATE $k^* \gets \text{argmin}_k L_k$
\STATE \textbf{return} $\theta_{k^*}$
\STATE \color{teal} \# Report $L = $ Error rate of $w_{k^*}$ on $D_{test}$. \color{black}
\end{algorithmic}
\end{algorithm}

\pagebreak
\section{Random Search Pseudocode}
\label{sec:appendix:pseudocode}
We show examples of how RS is used to choose HPs in centralized vs. federated learning in Algorithms~\ref{alg:rs} and~\ref{alg:rs_fl}. To generally adapt traditional HP tuning methods (e.g. RS, TPE, HB, and BOHB) to FL, we simply replace the original training / evaluation subroutines with federated versions.  These subroutines are highlighted in red in the RS example.

\begin{algorithm}
\caption{RS for subsampled FL clients}\label{alg:rs_fl}
\begin{algorithmic}
\REQUIRE 
$\Theta$ (hyperparameter space)
\\ $K$ (num. of HP configs to search)
\\ $R$ (training rounds per HP config).
\\ $s_{\text{tr}}$ (num. of clients sampled per train round)
\\ $s_{\text{val}}$ (num. of clients sampled per eval round)
\\ $N_{\text{tr}}$ (total num. of training clients)
\\ $N_{\text{val}}$ (total num. of validation clients)
\\ $\{D_{\text{tr},i}\}_{i=1}^{N_{\text{tr}}}$ (training clients' data)
\\ $\{D_{\text{val},i}\}_{i=1}^{N_{\text{val}}}$ (validation clients' data)
\\ $\{p_{\text{val},i}\}_{i=1}^{N_{\text{val}}}$ (validation clients' weights)
\\ \texttt{ServerOPT}() (model aggregation method)
\\ \texttt{ClientOPT}() (local optimization method)
\color{black}
\FOR {$k = 1, ..., K$}
    \STATE Sample HP config $\theta_k\sim\Theta$ uniformly at random.
    \STATE Initialize model parameters $w_k$.
    \STATE \color{teal} \textbf{\# Federated Training} \color{black}
    \FOR {$r = 1, ..., R$}        
    \color{red}
        \STATE Sample clients $
        a_1, ..., a_{s_\text{tr}}$ where $a_i\sim[N_{\text{tr}}]$ is sampled uniformly without replacement.
        \FOR{$i=1,...,s_{\text{tr}}$}
            \STATE $w'_{a_i} \gets$ \texttt{ClientOPT}($w_k, D_{\text{tr},a_i}; \theta_k$)
        \ENDFOR{}
        \STATE $w_k \gets \texttt{ServerOPT}(w_k, \{w'_{a_i}\}_{i=1}^{s_{\text{tr}}}; \theta_k)$
        \color{black}
    \ENDFOR
    \\ \color{teal} \textbf{\# Federated Evaluation}
    \color{red}
    \STATE Sample clients $a_1, ..., a_{s_\text{val}}$ where $a_j\sim[N_{\text{val}}]$ is sampled uniformly without replacement.
    \FOR{$j=1,...,s_{\text{val}}$}
        \STATE $F_{\text{val},a_k} \gets$ Error rate of $w_k$ on client data $D_{\text{val},a_j}$
    \ENDFOR
    \STATE $L_k = (\sum_{j=1}^{s_{\text{val}}} p_{\text{val},a_j}F_{\text{val},a_j}) / \sum_{j=1}^{s_{\text{val}}} p_{\text{val},a_j}$ (Eq.~\ref{eq:fl_eval})
\color{black}
\ENDFOR
\STATE $k^* \gets \text{argmin}_k L_k$
\STATE \textbf{return} $\theta_{k^*}$
\STATE \color{teal} \# Report the full validation error rate:
\FOR{$j=1,...,N_{\text{val}}$}
    \STATE $F_{\text{val},a_j} \gets$ Error rate of $w_{k^*}$ on client data $D_{\text{val},a_j}$
\ENDFOR
\STATE Report $L = (\sum_{j=1}^{N_{\text{val}}} p_{\text{val},a_j}F_{\text{val},a_j}) / \sum_{j=1}^{N_{\text{val}}} p_{\text{val},a_j}$
\end{algorithmic}
\end{algorithm}

\begin{table*}
\centering
    \begin{tabular}{lccccccc}
        \hline
        & & \multicolumn{2}{c}{Clients} & \multicolumn{3}{c}{\# Examples (images/sequences)} \\
        Dataset & Task & Train & Eval & Mean & Min & Max & Total \\
        \hline
        CIFAR10 & Image Classification & 400 & 100 & 100 & 83 & 131 & 5K\\
        FEMNIST & Image Classification & 3,507 & 360 & 203 & 19 & 393 & 73K \\
  StackOverflow & Next Token Prediction & 10,815 & 3,678 & 391 & 1 & 194,167 & 5.6M \\
        Reddit  & Next Token Prediction & 40,000 & 9,928 & 19 & 1 & 14,440 & 1.1M \\
         
    \end{tabular}
    \caption{More detailed dataset statistics.}
    \label{tab:appendix:datasets}
\end{table*}

\pagebreak
\clearpage
%
%
%
%
%





\section{Artifact Appendix}

\subsection{Abstract}

We provide several \texttt{Python} scripts to tune FedAdam and a \texttt{Jupyter Notebook} to analyze the results. A machine with a single \texttt{CUDA}-supported GPU and 4-core CPU is sufficient to validate results on CIFAR10 / FEMNIST. We recommend using multiple GPUs to run trials of StackOverflow / Reddit in parallel.

\subsection{Artifact check-list (meta-information)}

{\small
\begin{itemize}
  \item {\bf Algorithm:} Random search, Tree-structured Parzen Estimator, Hyperband, BOHB, FedAdam
  \item {\bf Data set:} CIFAR10, FEMNIST, StackOverflow, Reddit
  \item {\bf Hardware:} NVIDIA GeForce GTX 1080 Ti
  \item {\bf How much disk space required (approximately)?:} 50GB
  \item {\bf How much time is needed to prepare workflow (approximately)?:} 1 hour
  \item {\bf How much time is needed to complete experiments (approximately)?:} 1000 GPU hours (full experiments). 1 hour (analysis only).
  \item {\bf Publicly available?:} Yes
  \item {\bf Code licenses (if publicly available)?:}  Apache License 2.0
  \item {\bf Data licenses (if publicly available)?:} BSD-2-Clause license (LEAF), Creative Commons Attribution-ShareAlike 3.0 Unported License (StackOverflow)
  \item {\bf Workflow framework used?:} VSCode
  \item {\bf Archived (provide DOI)?: 10.48550/arXiv.2212.08930
}
\end{itemize}

\subsection{Description}

\subsubsection{How delivered}

The artifacts and step-by-step experiment instructions are located at the Github repository: \url{https://github.com/imkevinkuo/noisy-eval-in-fl}. We additionally provide a copy of the artifacts at Zenodo \citep{kevin_kuo_2023_7819606}.

\subsubsection{Hardware dependencies}

The experiments require 4 to 12GB of memory (4GB for image, 12GB for text) on a \texttt{CUDA}-enabled GPU and 4GB of memory on the host machine. 50GB of disk space is needed to store the datasets and results. 
The GPU runtime is approximately split 25 / 25 / 650 / 300 hours across CIFAR10 / FEMNIST / StackOverflow / Reddit respectively.

\subsubsection{Software dependencies}

All code is written in \texttt{Python} (3.9.12). The critical \texttt{Python} libraries required for training are \texttt{PyTorch} (1.11.0) and \texttt{Numpy} (1.22.3). Additionally, we use \texttt{CUDA} (11.6) which allows \texttt{PyTorch} to perform tensor operations on \texttt{CUDA}-enabled GPUs. A complete list of package requirements can be found in the Github repository's \texttt{environment.yml}.

\subsubsection{Data sets}

\subsection{Installation}

To set up the code, pull the Github repository and follow the instructions in \texttt{README.md}. We will provide pre-processed versions of the datasets which can be downloaded within an hour. Otherwise, setting up the datasets from scratch can take up to 5 hours.

\subsection{Experiment workflow}

The main scripts have a prefix of \texttt{fedtrain\_*.py} and have a suffix of either \texttt{simple}, \texttt{bohb}, or \texttt{tpe}. \texttt{simple} trains a single model for a given FedAdam HP configuration. These runs are used in \texttt{analysis.ipynb} to simulate the outcome of RS and HB. \texttt{bohb} and \texttt{tpe} run the respective HP tuning algorithms and depend on \texttt{simple} for model training and evaluation.

A set of helper scripts have a prefix of \texttt{init\_*.py}. Each \texttt{init} script is a wrapper which runs multiple trials of the corresponding \texttt{fedtrain} script. To complete the training portion of the experiments, run each of \texttt{init} scripts once. The number of trials can lowered within the \texttt{init} files.

After training the models, plots can be generated by running all the cells in \texttt{analysis.ipynb}.

\subsection{Evaluation and expected result}

We briefly describe the expected results which correspond to each major observation we make in the main paper:
\begin{enumerate}
    \item \textbf{(Subsampling)} The curves should trend towards a lower error rate as the number of subsampled clients increases. \texttt{Best HPs} should be a horizontal line below each curve.
    \item \textbf{(Budget)} The curves should trend towards a lower error rate as training rounds increases. There should be a noticable gap between the 1 client and 100\% client curves.
    \item \textbf{(Data Heterogeneity)} Curves with niid data ($p=1$) should have a higher error rate than those with iid data ($p=0$).
    \item \textbf{(Systems Heterogeneity)} On CIFAR10 and Reddit, curves with a larger value of $b$ should have a larger error rate.
    \item \textbf{(Privacy)} Curves with a smaller value of $\eps$ should have a larger error rate.
    \item \textbf{(HPO Degradation)} RS and TPE should degrade less than HB and BOHB do when applying subsampling ($1\%$) and DP evaluation ($\eps=100$).
    \item \textbf{(Proxy Data)} The scatter plots for CIFAR10/FEMNIST and StackOverflow/Reddit should show a positive correlation between a configuration's error rate on the two datasets.
    \item \textbf{(Proxy Data vs. Noisy Eval)} Tuning with the best proxy dataset should outperform tuning with subsampling ($1\%$) and DP evaluation ($\eps=1$).

\end{enumerate}



\subsection{Methodology}

Submission, reviewing and badging methodology:

\begin{itemize}
  \item \url{http://cTuning.org/ae/submission-20190109.html}
  \item \url{http://cTuning.org/ae/reviewing-20190109.html}
  \item \url{https://www.acm.org/publications/policies/artifact-review-badging}
\end{itemize}



\end{document}